\documentclass[final,5p,times,twocolumn,authoryear]{elsarticle}

\usepackage{amssymb}

\usepackage{hyperref}
\usepackage{epsfig}
\usepackage{amsmath}
\usepackage{xcolor}
\usepackage[caption=false, font=footnotesize, position=top]{subfig}

\journal{Computers and Electronics in Agriculture}

\begin{document}

\begin{frontmatter}

\title{Overcoming Small Minirhizotron Datasets Using Transfer Learning}

\author[1]{Weihuang Xu}
\ead{weihuang.xu@ufl.edu}

\author[1]{Guohao Yu}
\ead{guohaoyu@ufl.edu}

\author[1]{Alina Zare}
\ead{azare@ufl.edu}

\author[2]{Brendan Zurweller}
\ead{brendan.zurweller@msstate.edu}

\author[3]{Diane Rowland}
\ead{dlrowland@ufl.edu}

\author[4]{Joel Reyes-Cabrera}
\ead{reyescabreraj@missouri.edu}

\author[4]{Felix B. Fritschi}
\ead{fritschif@missouri.edu}

\author[5]{Roser Matamala}
\ead{matamala@anl.gov}

\author[6]{Thomas E. Juenger}
\ead{tjuenger@austin.utexas.edu}

\address[1]{Department of Electrical and Computer Engineering, University of Florida, FL, USA}
\address[2]{Department of Plant and Soil Science, Mississippi State University, MS, USA}
\address[3]{Department of Agronomy, University of Florida, FL, USA}
\address[4]{Division of Plant Sciences, University of Missouri, MO, USA}
\address[5]{Argonne National Laboratory, IL, USA}
\address[6]{Department of Integrative Biology, University of Texas at Austin, TX, USA}


\begin{abstract}
Minirhizotron technology is widely used to study root growth and development. Yet, standard approaches for tracing roots in minirhiztron imagery is extremely tedious and time consuming.  Machine learning approaches can help to automate this task. However, lack of enough annotated training data is a major limitation for the application of machine learning methods.
Transfer learning is a useful technique to help with training when available datasets are limited.
In this paper, we investigated the effect of pre-trained features from the massive-scale, irrelevant ImageNet dataset and a relatively moderate-scale, but relevant peanut root dataset on switchgrass root imagery segmentation applications. 
We compiled two minirhizotron image datasets to accomplish this study: one with 17,550 peanut root images and another with 28 switchgrass root images. 
Both datasets were paired with manually labeled ground truth masks.
Deep neural networks based on the U-net architecture were used with different pre-trained features as initialization for automated, precise pixel-wise root segmentation in minirhizotron imaghursery.
We observed that features pre-trained on a closely related but relatively moderate size dataset like our peanut dataset were more effective than features pre-trained on the large but unrelated ImageNet dataset.
We achieved high quality segmentation on peanut root dataset with 99.04\% accuracy at the pixel-level and overcame errors in human-labeled ground truth masks.
By applying transfer learning technique on limited switchgrass dataset with features pre-trained on peanut dataset, we obtained 99\% segmentation accuracy in switchgrass imagery using only 21 images for training (fine tuning).
Furthermore, the peanut pre-trained features can help the model converge faster and have much more stable performance.
We presented a demo of plant root segmentation for all models under \url{https://github.com/GatorSense/PlantRootSeg}.
\end{abstract}

\begin{keyword}
Deep learning \sep Transfer learning \sep Root and soil segmentation \sep Plant root phenotyping \sep Minirhizotron

\end{keyword}

\end{frontmatter}


\section{Introduction}
Minirhizotron camera systems are a minimally-invasive imaging technology for monitoring and understanding the development of plant root systems\citep{majdi1996root}. 
Such systems collect visible-wavelength color imagery of plant roots in-situ by scanning an imaging system within a clear tube driven into the soil.
A time-series of root images can be captured by the Minirhizotron camera systems to record the changes of each individual root such as birth, death, elongation, etc. 
The time-series analysis of these features are crucial to understand the ecosystem, which is carried out by powerful tools such as rhizoTrak \citep{moller2019rhizotrak}.
A variety of root phenotypes can be determined from minirhizotron RGB root imagery, such as lengths, diameters, patterns, turnover and distributions at different depths.
Automated analysis of root systems may facilitate new scientific discoveries that could be applied to address the world's pressing food, resource, and climate issues.
A key component of automated analysis of plant roots from imagery is the pixel-level segmentation of roots from their surrounding soil.
However, manually tracing roots in minirhizotron imagery is tedious and extremely time-consuming, which limits the number and size of experiments.
Thus, techniques that can automatically and accurately segment roots from minirhizotron imagery are crucial to improve the efficiency of data collection and post-processing.
Many methods in the literature based on both traditional image processing and learning methods \citep{zeng2006detecting, zeng2010rapid, shojaedini2013new} and  deep learning methods \citep{yasrab2019rootnav, wang2019segroot} have been developed to tackle this problem.
However, the performance of these previous methods are usually limited by the quantity and quality of the data, especially for the approaches based on deep learning.

Root segmentation belongs to the field of semantic image segmentation, which is one of the most challenging tasks in computer vision.
Instead of assigning labels at the whole-image level for image classification problems, semantic image segmentation requires a model to predict a label for each pixel.
Many methods based on deep convolutional neural networks (DCNN) have been proposed to address semantic segmentation tasks such as fully convolutional networks\citep{long2015fully}, SegNet\citep{badrinarayanan2015segnet}, U-net\citep{ronneberger2015u}, and DeepLab\citep{chen2018deeplab}. 
Models based on the above methods have achieved success in segmentation of medical images\citep{ronneberger2015u,pandey2018segmentation,chen2017dcan,zhang2019prostate}, satellite images\citep{constantin2018accurate,rakhlin2018land,bai2018towards}, and plant images\citep{chen2018automatic,zhu2018data}. 
Such segmentation models are based on supervised learning of large networks with a very large number of parameters, requiring a huge amount of data with ground truth to achieve a satisfactory performance. 
Deep neural networks trained on small datasets can quickly overfit for the small sets and perform poorly in larger unknown sets.
Thus, a fundamental issue of using those models for many applications, including plant science, is limited availability of training data.

To address such problems, so-called transfer learning\citep{girshick2014rich, bengio2012deep} techniques have been developed that apply model-weights pre-trained on large-scale data as initial parameters, and then fine-tune the models on target problems that usually have more limited training data. 
This process will work based on the assumption that those pre-trained features are fairly general and applicable to many visual image applications, and can be re-used for a different specific problem. When the target dataset is small, pre-trained features can significantly improve the performance and help with faster convergence. Leveraging this idea, features pre-trained on massive scale data such as ImageNet are widely used as initial weights in recent work, which achieved state-of-the-art results on a variety of different tasks, such as image classification\citep{donahue2014decaf, sharif2014cnn}, object detection\citep{girshick2014rich, ren2015faster,sermanet2013overfeat} and image segmentation\citep{iglovikov2018ternausnet, chen2018automatic, dai2016instance}. The ImageNet dataset\citep{deng2009imagenet} contains more than 14 million pictures which is divided into 1000 classes. Classes are varied; example classes include images of sea lions, volcanos, chain saws, and others.  Each image is quality-controlled and human-annotated. Great breakthroughs have been achieved based on ImageNet dataset in different research areas such as image classification, object localization, object detection, scene classification and Scene parsing.
However, more and more work is questioning the effects of ImageNet pre-training. Huh et al. \citep{huh2016makes} illustrated that transfer learning performance is similar with features pre-trained only on half of ImageNet dataset as opposed to the full dataset. More relevant to this study, Yosinski et al. \citep{yosinski2014transferable} showed that features in shallow layers are more general and effective when transferred to other specific problems. On the contrary, features from higher layers are often more problem-specific. 
Thus, we suppose that ImageNet pre-training is less effective as compared to discipline-specific data sets that include plant root imagery, due to the lack of relevance of the ImageNet dataset relative to plant root datasets.

In this work, we collected a moderately sized peanut root minirhizotron imagery dataset and a small sized switchgrass root minirhizotron imagery dataset and manually traced root segments in both sets.
We trained U-net based models with different depths on the peanut root dataset to achieve automated, precise pixel-wise root segmentation. 
We also investigated and compared the effect on segmentation performance of model depth to find an appropriate model depth for further transfer learning study. Then, we used a transfer learning approach to apply pre-trained features from the peanut root data and the ImageNet dataset on the small-scale switchgrass root dataset to explore the effect of different pre-trained features and the accuracy of segmentation. 
Our results affirm that features pre-trained on a moderate-sized peanut dataset that was highly related to the target switchgrass dataset were more effective for our root segmentation problem than those pre-trained on the large-scale but less relevant ImageNet dataset. Furthermore, our results also confirm that pre-training via transfer learning techniques is effective (and necessary) for small-scale plant root datasets.  
Beyond this, our results also indicate the importance of applying the pre-trained features of full network (as opposed to only encoder features) to ensure more reliability of results both in accuracy and stability (lower variance across different train-test trials).

In the following sections we describe our datasets, the semantic segmentation methods employed, our experiments using those methods with our datasets and other visual imagery datasets, and finally draw some conclusions based upon that work.

\section{Datasets}
We have compiled two minirhizotron root image datasets. 
The first dataset contains 17,550 peanut root RGB images and the second dataset has 28 switchgrass root RGB images.
The pixel size of the peanut and switchgrass imagery are 760 x 580 and 2160 x 1560, respectively.
All images in both datasets were acquired using minirhizotron systems in the field, and were paired with manually labeled ground truth masks indicating the location of roots in each image. 
The details of data collection and labelling process are as follows.

\begin{figure}[h]
\begin{center}
  \subfloat[Raw Peanut Root Image]{%
       \includegraphics[width=0.48\linewidth]{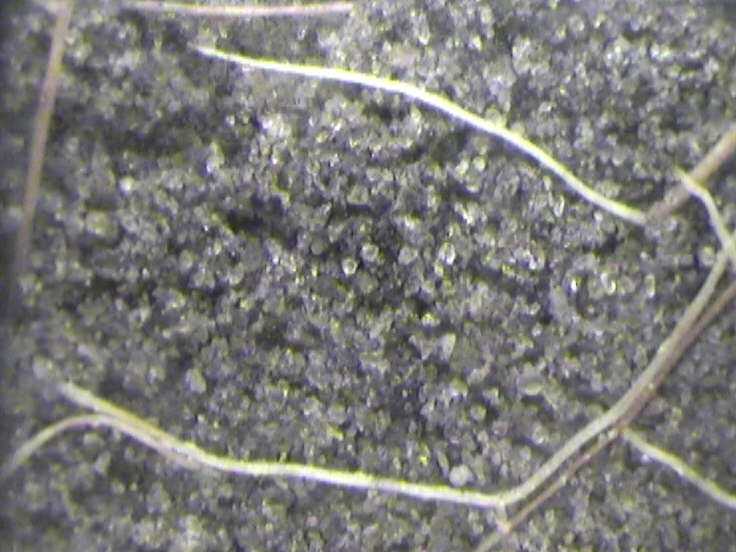}}
    \hfill
  \subfloat[Manually Labeled GT]{%
        \includegraphics[width=0.48\linewidth]{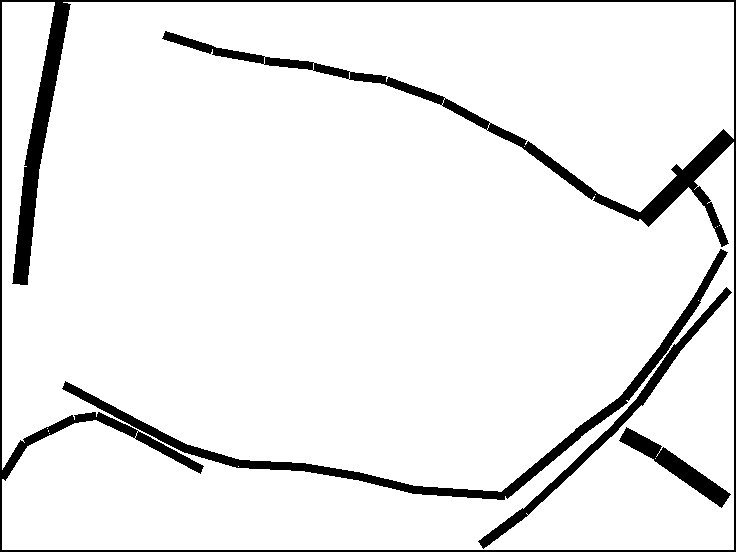}}
    \\
  \subfloat{%
        \includegraphics[width=0.48\linewidth]{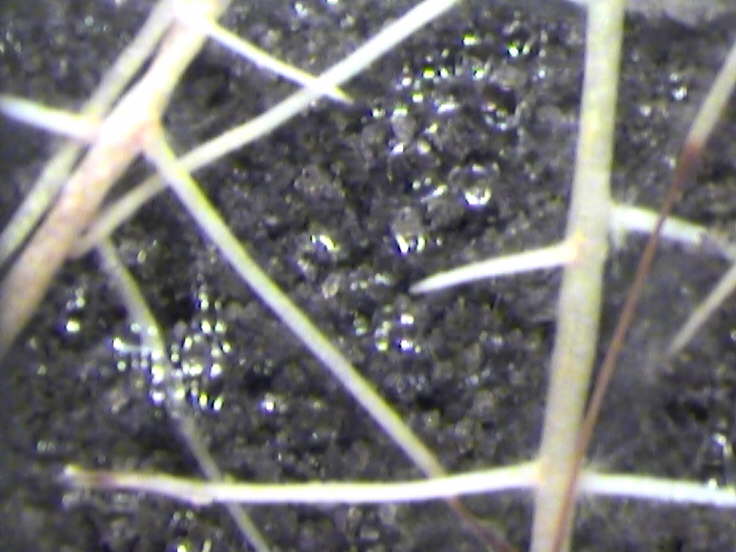}}
    \hfill
  \subfloat{%
        \includegraphics[width=0.48\linewidth]{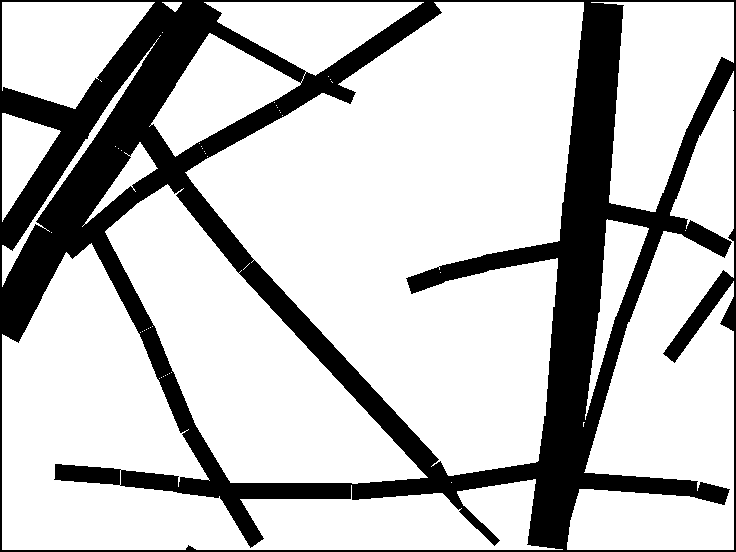}}
    \\
  \subfloat{%
       \includegraphics[width=0.48\linewidth]{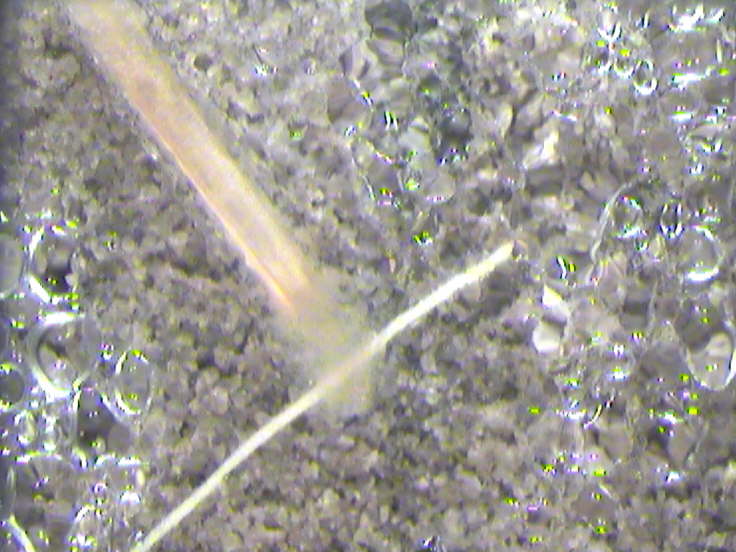}}
    \hfill
  \subfloat{%
        \includegraphics[width=0.48\linewidth]{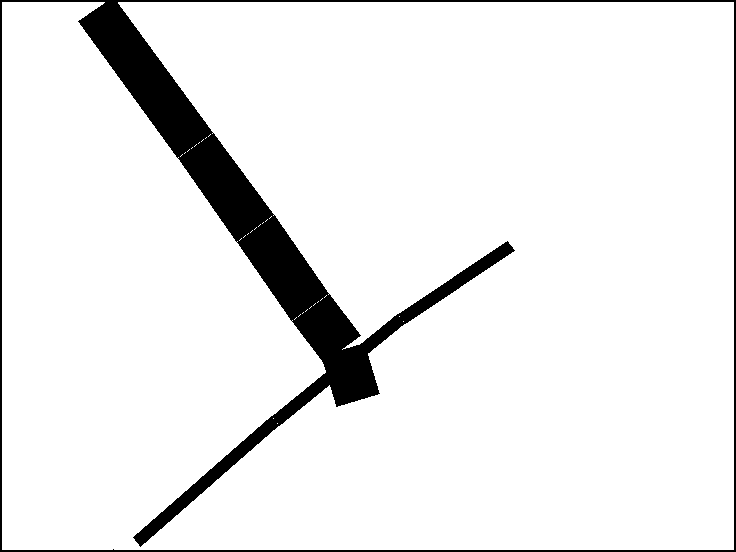}}
    \\
  \subfloat{%
        \includegraphics[width=0.48\linewidth]{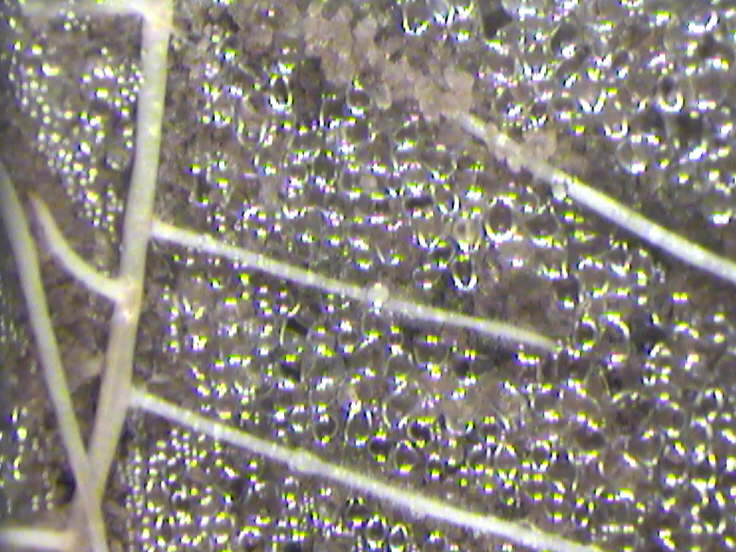}}
    \hfill
  \subfloat{%
        \includegraphics[width=0.48\linewidth]{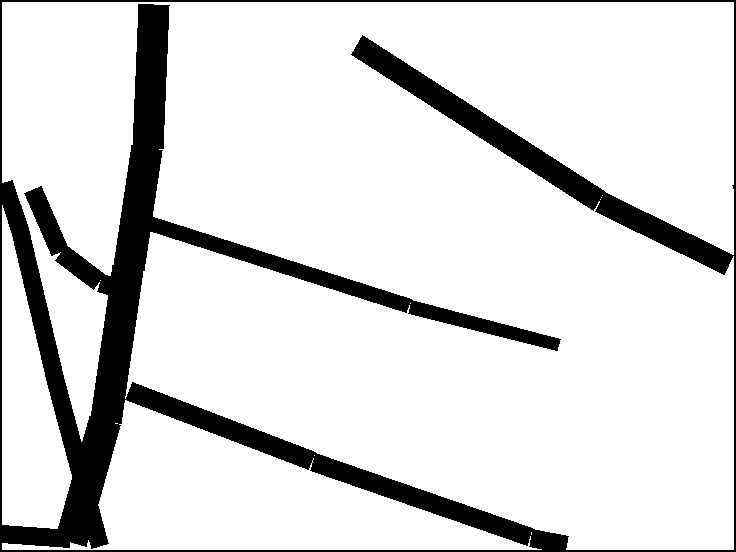}}
\end{center}
  \caption{Examples of peanut root minirhizotron images (column (a)) captured by minirhizotron camera systems and manually labeled ground truth masks (column (b)) generated using WinRHIZO software.}
  \label{Peanut} 
\end{figure}

\textbf{Peanut root dataset} was collected in a field trial at the Plant Science Research and Education Unit (PSREU) located in Citra, Florida during the 2016 growing season. Minirhizotron tubes 2 m in length were installed directly under and parallel to the row at a $45^{\circ}$ angle to the soil surface after crop emergence using a hydraulic powered coring machine (Giddings Machine Company, Windsor, CO). 
After installation, the portion of the minirhizotron tube protruding from the soil was covered with reflectance insulation (Reflectix Inc., Markleville, IN) to avoid root light exposure and precipitation from entering the tube. At each measurement date, images were captured at 150 dots per inch (5.9 pixels per millimeter) at 13.5 mm increments (resulting typically in 112 image frames) along the minirhizotron tubes using a BTC 100X video camera and BTC I-CAP image capture software (Bartz Technology Corporation, Carpinteria, CA). Root parameter analysis was conducted using WinRHIZO Tron software (Regent Instruments Inc., Quebec, Canada) by hand tracing root segments within each image frame. The binary ground truth masks were generated by hand using the WinRHIZO Tron software package. The process consists of manually drawing different sizes of rectangles to highlight the area of roots while attempting to leave the soil blank. Examples of collected peanut root images and corresponding labeled ground truth masks are shown in Figure \ref{Peanut}. Labelling by WinRHIZO is faster than labelling images pixel by pixel, since a large area of root pixels can be labeled at once. However, two shortcomings of this method are that: 1) the width of each rectangle is constant indicating that the roots in the labeled area have the same diameter, which is not true in practice; and 2) when the roots are not straight, there are gaps between labeled regions and the labeled edges of roots are not smooth. 

\begin{figure}
\begin{center}
  \subfloat[Raw Switchgrass Root Image]{%
      \includegraphics[width=0.48\linewidth]{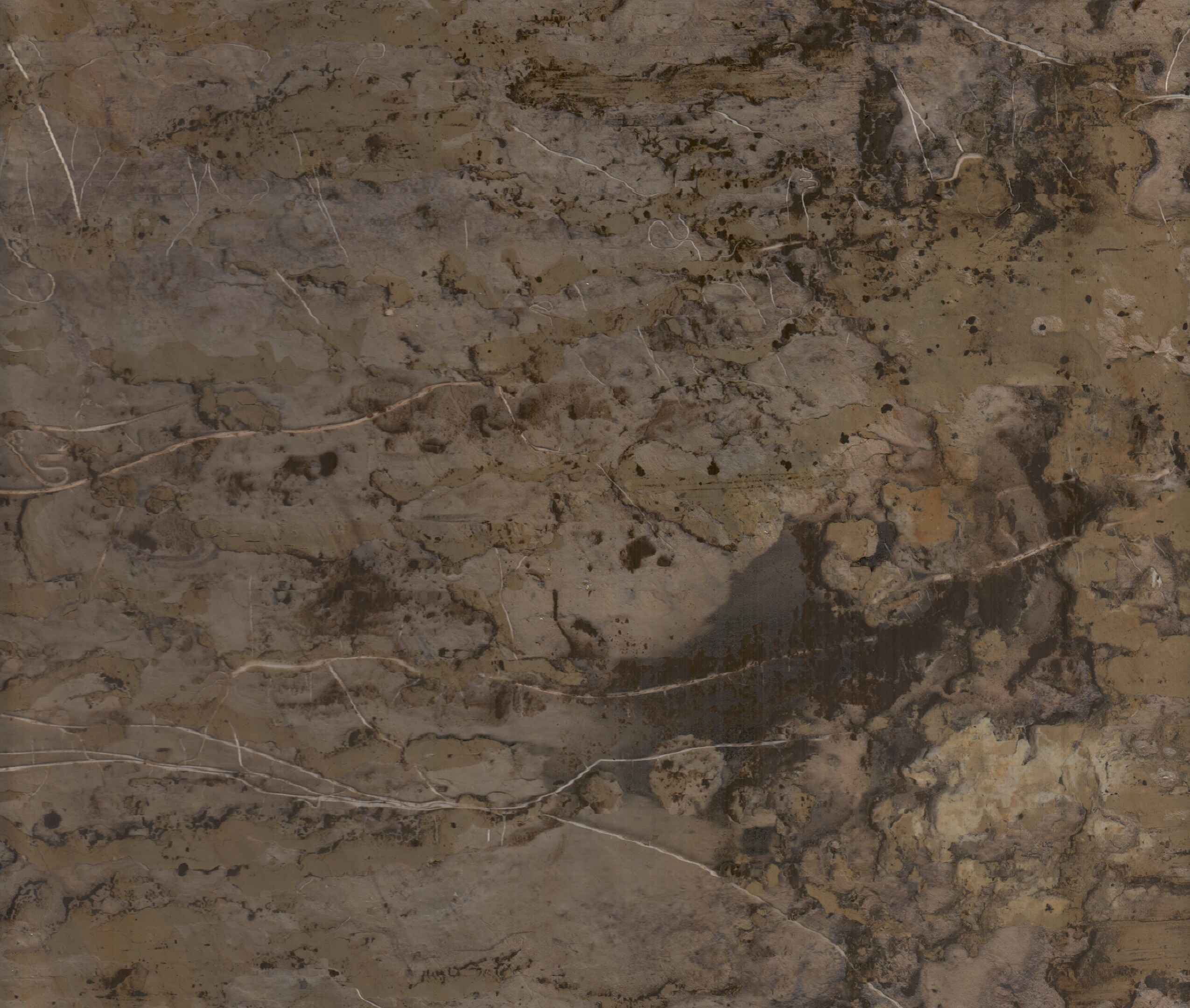}}
    \hfill
  \subfloat[Manually Labeled GT]{%
      \includegraphics[width=0.48\linewidth]{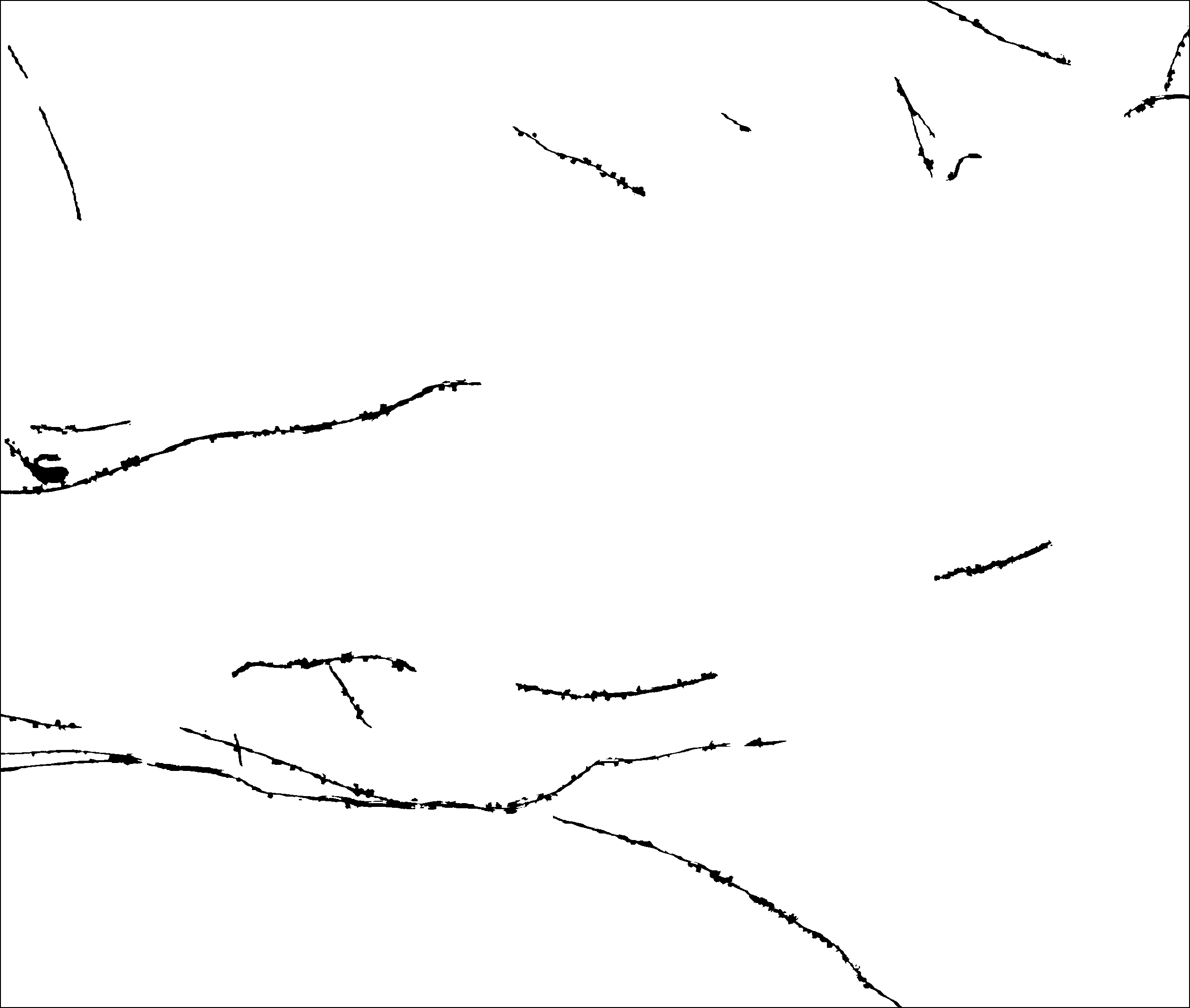}}
    \hfill
    \\
  \subfloat{%
        \includegraphics[width=0.48\linewidth]{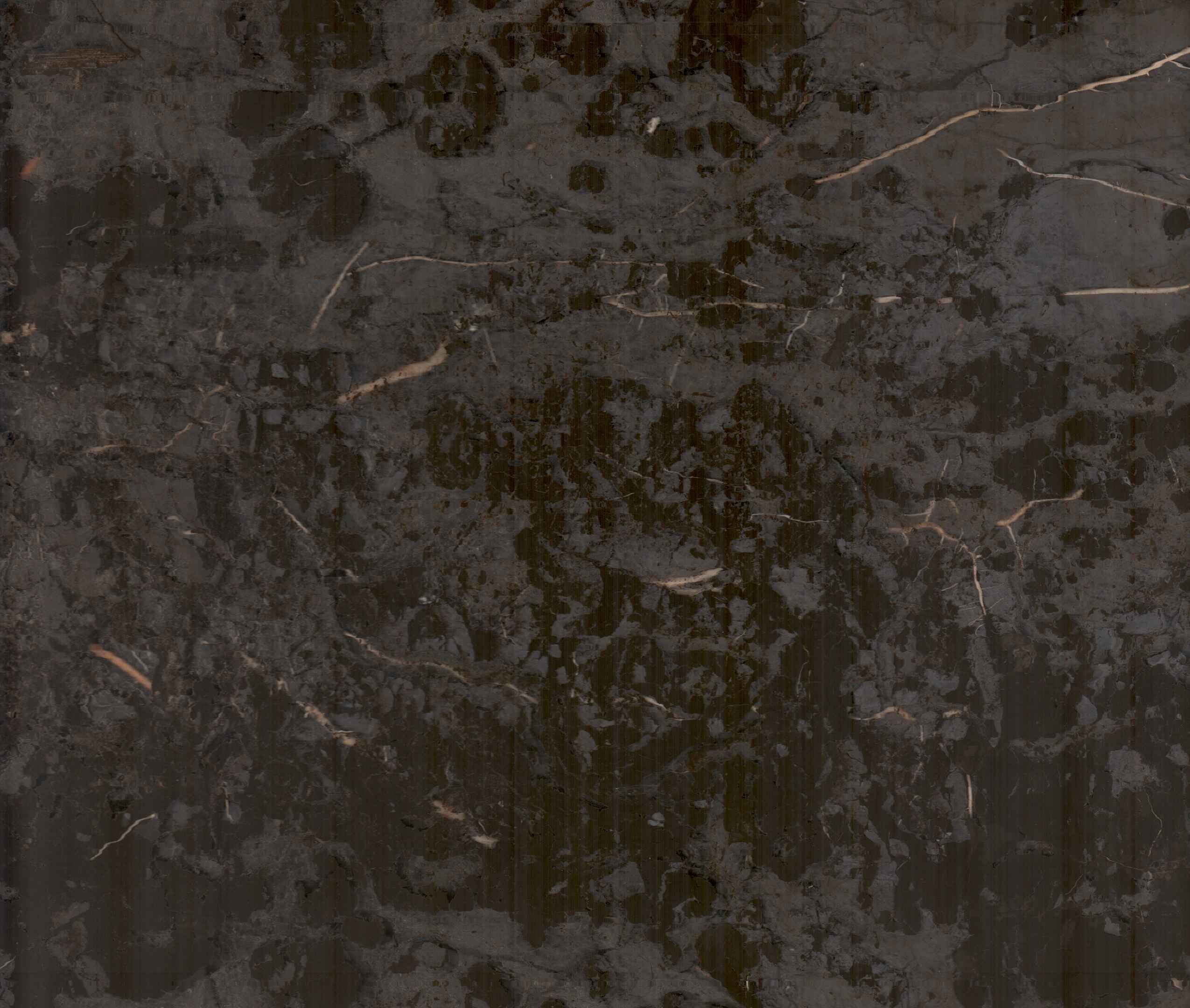}}
    \hfill
  \subfloat{%
        \includegraphics[width=0.48\linewidth]{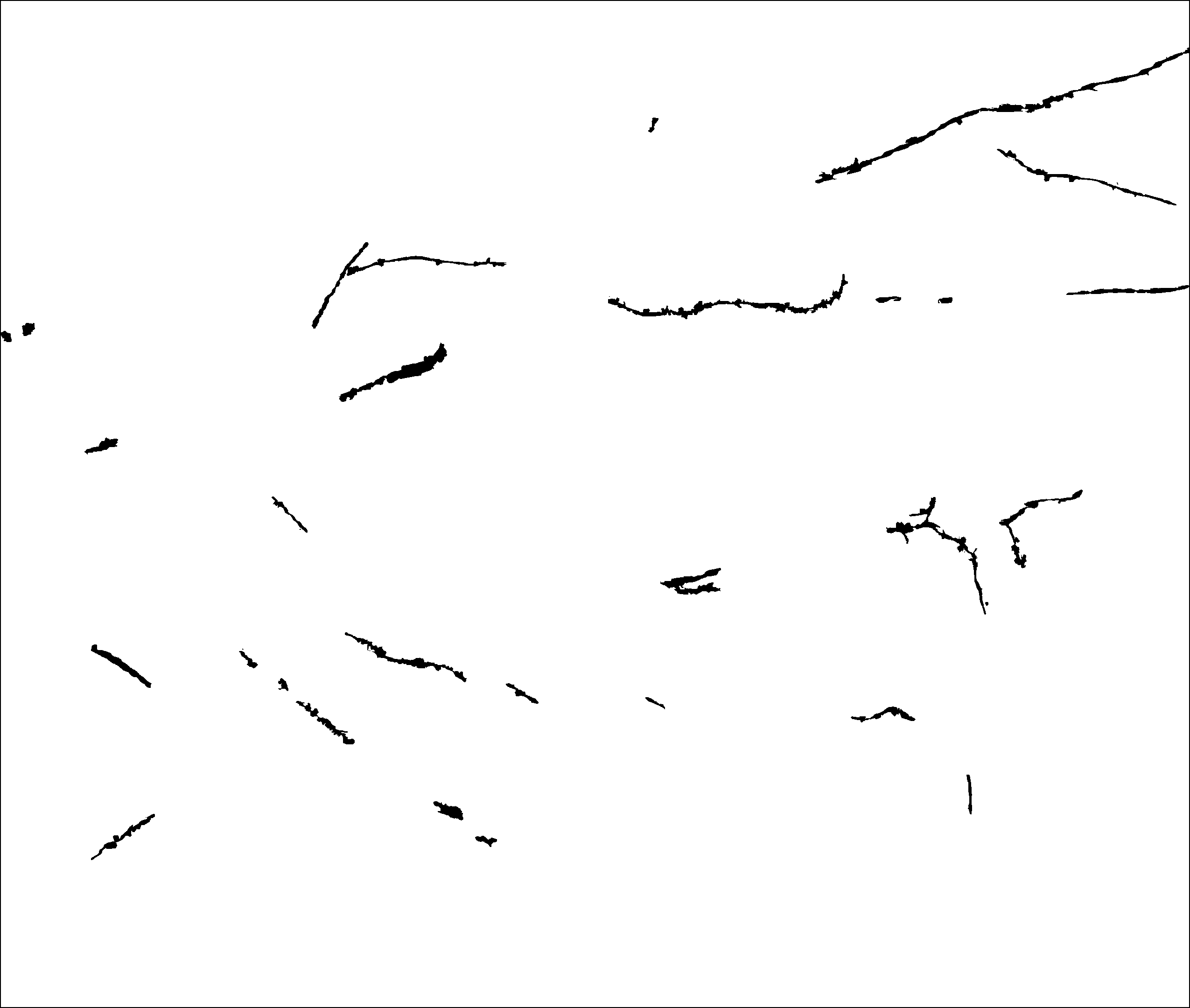}}
\end{center}
  \caption{Examples of switchgrass root images (column (a)) captured by minirhizotron camera  systems and manually labeled ground truth masks (column (b)).}
  \label{Switchgrass} 
\end{figure}

\textbf{Switchgrass root dataset} was collected using a CI-602 in-situ root imager (CID Bio- Science, Camas, WA, USA) in minirhizotron tubes in a 2-year old switchgrass field at the U.S. Department of Energy National Environmental Research Park at Fermilab in Batavia, IL, USA. Minirhizotron tubes were installed with an angle of $60^{\circ}$ to the soil surface to an approximate maximum vertical depth of 120 cm using an angled guided soil core sampler. Foam caps were installed to eliminate root light exposure, reduce temperature variation in the root zone, and stop precipitation from entering the tubes. Switchgrass root images were taken at 300 dots per inch (11.8 pixels per millimeter) from eight depth intervals along the minirhizotron tubes. We ran the Simple Linear Iterative Clustering (SLIC) algorithm \citep{achanta2010slic} on raw root images to cluster spatially-contiguous pixels into groups according to color (pixel values). Then, the root images were pre-segmented into superpixel level where each superpixel contained around 100 pixels in raw images. All the binary ground truth masks were manually labeled on superpixel level, which improved efficiency of labelling process. All the binary ground truth masks were manually labeled on superpixel level \citep{yu2019root}. Examples of raw switchgrass images and corresponding ground truth masks are shown in Figure \ref{Switchgrass}.

\begin{figure*}[h]
\begin{center}
    {\includegraphics[width=2.0\columnwidth]{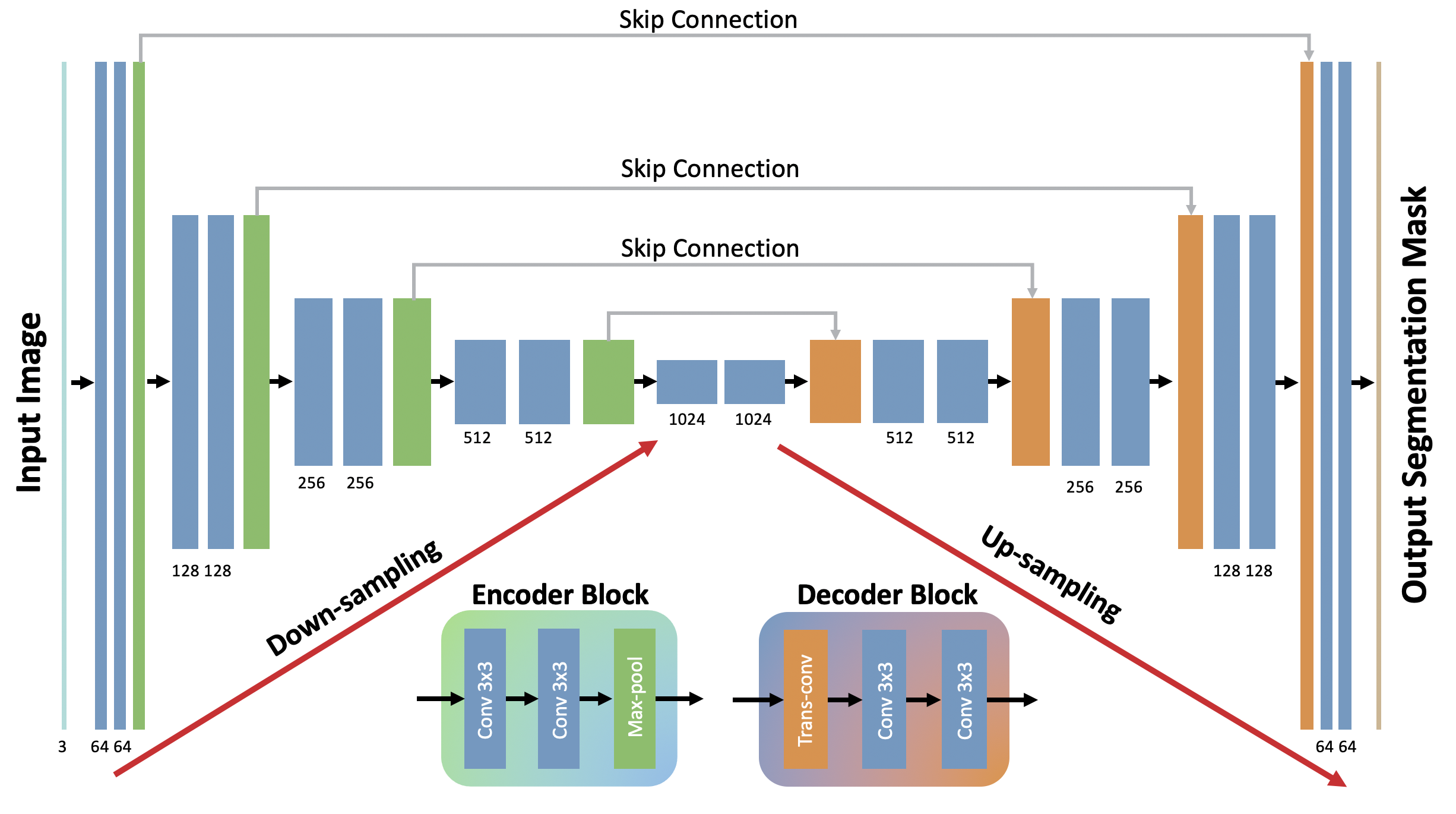}}
\end{center}
\caption{An illustration of the U-Net based encoder-decoder architecture. Each encoder block consists of two 3x3 convolution layers with ReLU (blue rectangle) and one 2x2 max-pooling layer (green rectangle). Each decoder block consists of one transpose convolution layer (orange rectangle) and two 3x3 convolution layers. The number of output channels of each convolution layer is denoted at the bottom of each blue rectangle. Feature maps extracted from encoders are concatenated to the corresponding decoder via skip connections. A 1x1 convolution layer is used at the end to reduce the number of channels.}
\label{fig:Unet}
\end{figure*}

\section{Methods}
\subsection{Network Architecture}
A U-Net \citep{ronneberger2015u} based encoder-decoder neural network was used for root segmentation.
The network architecture is shown in Figure \ref{fig:Unet}.
The left half of the architecture works as an encoder where each block consists of two 3x3 convolution layers followed by one 2x2 max-pooling layer to downsample feature maps. 
The right half of the architecture works as a decoder where each block consists of one transpose convolution layer and two 3x3 convolution layers. 
The transpose convolution layer up-samples the size of feature maps by two. 
The encoder blocks are trained to extract dense feature maps from minirhizotron RGB imagery. Via skip connections, those feature maps will be concatenated with higher-level ones in corresponding decoders to offer more spatial information in output mask. 
The last layer is a 1x1 convolution layer (i.e., a weighted sum across all feature layers) to convert feature maps to a heat map. 
Then, the softmax function is used to assign class labels to each pixel. 
As it uses fully convolutional network architecture \citep{long2015fully}, a U-Net can be trained end-to-end with input images of any size. 
In order to keep the dimension of the output segmentation mask to be the same as input images, zero padding is used in every convolution layer. The model was implemented using the Pytorch library(1.1.0) and trained on a GTX 1080TI GPU with 12 GB of RAM.

\subsection{Evaluation Metrics}
The neural network will generate probability score for each pixel, representing the likelihood of a pixel belonging to the root class. 
The quality of the binary segmentation mask is sensitive to probability thresholds which usually vary for different models.
In our experiments, we used the the receiver operating characteristic (ROC) curves and precision-recall (PR) curves to evaluate our segmentation results.
ROC curves show the tradeoff between the true positive rate (TPR) and the false positive rate (FPR) for different thresholds.  
The value of the area under the curve (AUC) is usually calculated from ROC curves to evaluate overall classification performance (both root pixels and non-root pixels) using different probability thresholds.
The perfect score of AUC in ROC curve is 1, which indicate 100\% classification accuracy on all pixels.
The PR curves show the tradeoff between the precision and recall for different thresholds.
Recall is the same as TPR indicating the percentage of correctly classified root pixels over all root pixels. 
Precision indicates the percentage of true root pixels over all the root pixels the model predicted.
A higher AUC value for a PR curve shows that the model is more likely to make accurate prediction on root pixels.
In our experiments, TPR (recall), FPR and precision values were calculated at the pixel-level by comparing the output predicted mask to the manually labeled ground truth.

\section{Experiments and Results}
We first trained our model on the peanut root dataset to investigate the segmentation performance on plant root minirhizotron imagery when moderate-sized dataset is available. 
Then, we designed experiments to demonstrate how the pre-trained features can help improve segmentation performance when a limited switchgrass dataset is used. 
We also compared the effect of features pre-trained on the well-known massive scale ImageNet dataset and features pre-trained on our peanut root dataset.

\subsection{Segmentation Performance on Peanut Root Dataset}

\textbf{Experiment setup}.
Model depth is crucial to the performance, especially when the training data is limited.
Deeper networks can extract higher-level features to improve segmentation performance, but it is easier to overfit on small training datasets.
To find the proper model depth for our peanut root dataset and future use on transfer learning experiment on switchgrass dataset, we implemented three models with depth 4, depth 5 and depth 6, where model depth refers to the number of encoders and decoders in the down-sampling and up-sampling path.
The peanut root images were collected across different dates, tubes and depths.
In order to ensure that the testing data and training data shared the same distribution, we randomly picked 90\% of the images from each date, tube and depth for training and the remaining 10\% for testing.
The inputs to the model were entire images (instead of stacking small randomly selected patches). 
The model was trained end-to-end from scratch using the SGD optimizer with learning rate 0.0001 and momentum 0.8.
We set the batch size to be two due to GPU memory limitation and used binary cross-entropy as loss function.
All models were trained for 100 epochs for five trials with different randomly initialized parameters.
We plotted ROC curves to evaluate overall segmentation performance for each model and calculated average AUC values and standard deviation of AUC values based on five trials.

\begin{figure*}[p]
\begin{center}
  \subfloat[Peanut Root Image]{%
      \includegraphics[width=0.19\linewidth]{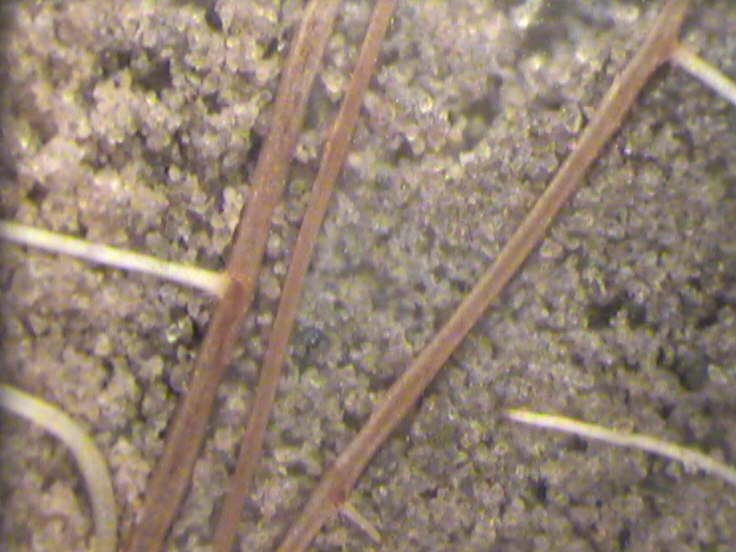}}
    \hfill
  \subfloat[Manually Labeled GT]{%
        \includegraphics[width=0.19\linewidth]{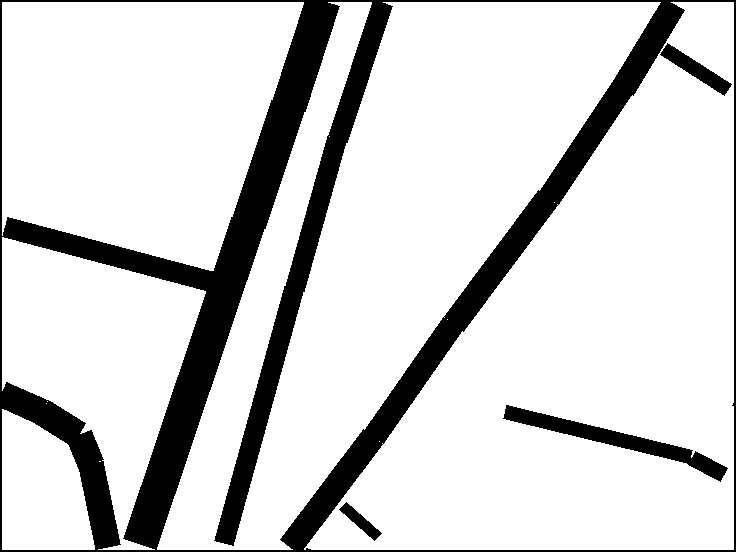}}
    \hfill
  \subfloat[Depth4 Model Output]{%
        \includegraphics[width=0.19\linewidth]{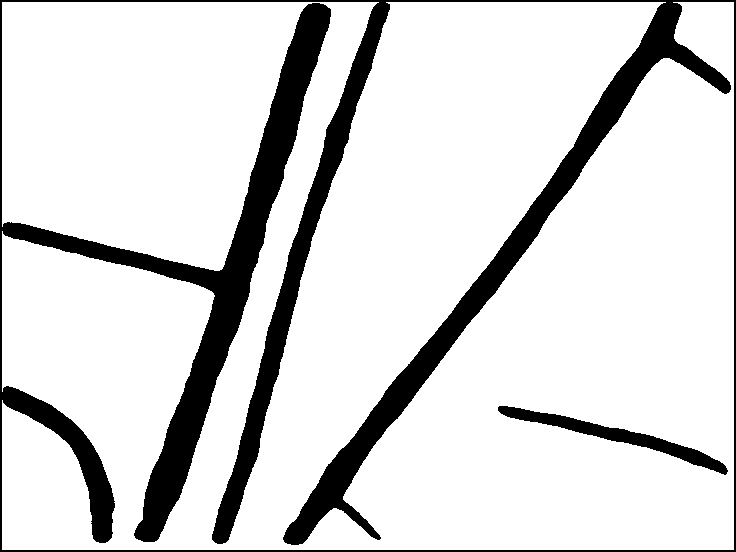}}
    \hfill
  \subfloat[Depth5 Model Output]{%
        \includegraphics[width=0.19\linewidth]{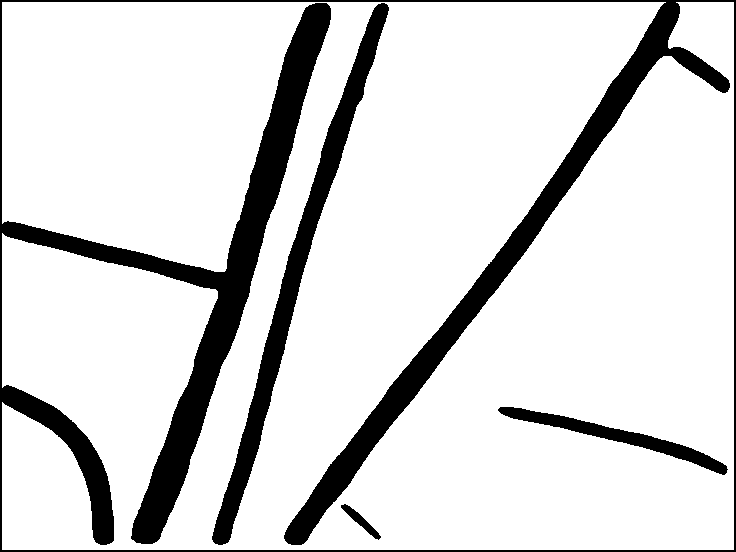}}
    \hfill
  \subfloat[Depth6 Model Output]{%
      \includegraphics[width=0.19\linewidth]{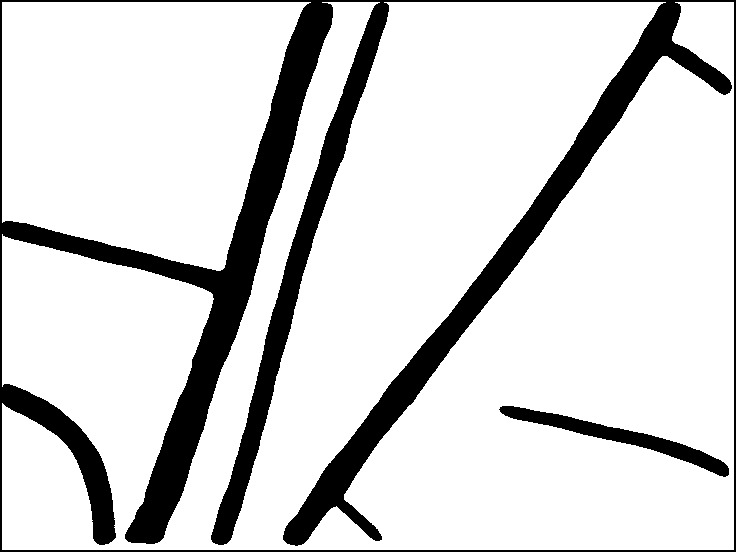}}
    \\
  \subfloat{%
        \includegraphics[width=0.19\linewidth]{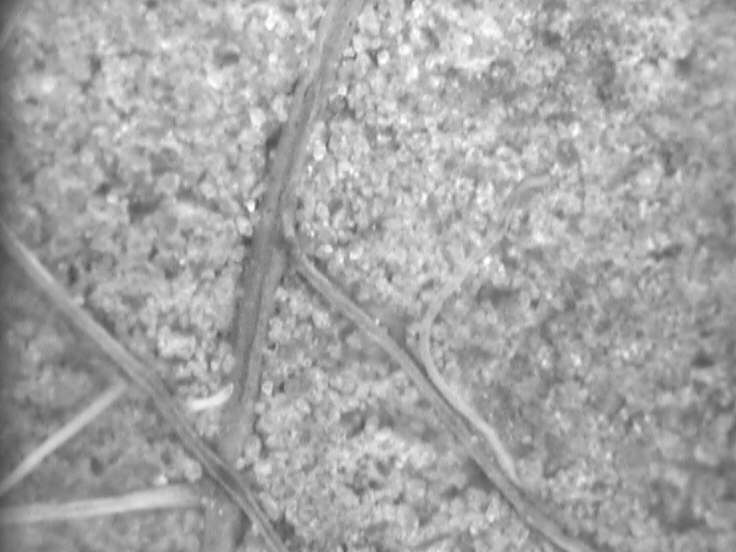}}
    \hfill
  \subfloat{%
        \includegraphics[width=0.19\linewidth]{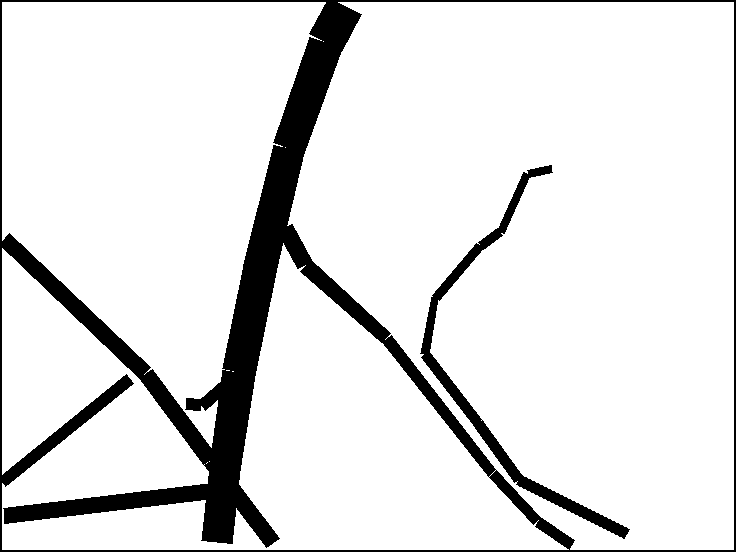}}
    \hfill
  \subfloat{%
        \includegraphics[width=0.19\linewidth]{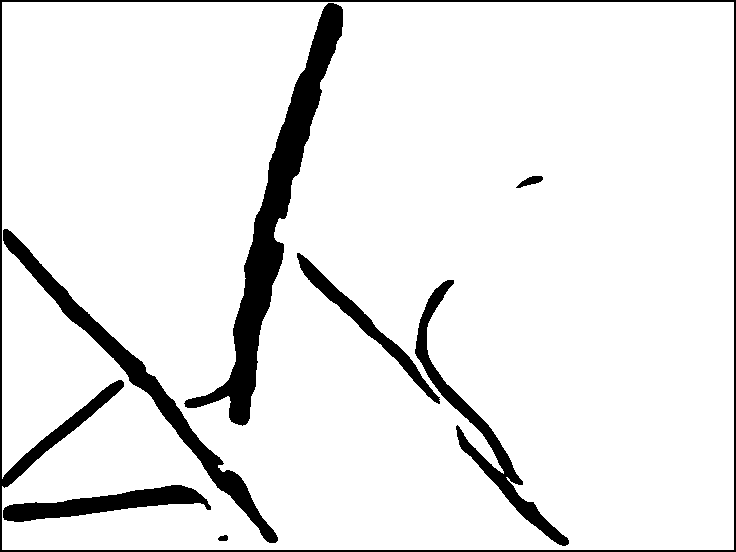}}
    \hfill
  \subfloat{%
        \includegraphics[width=0.19\linewidth]{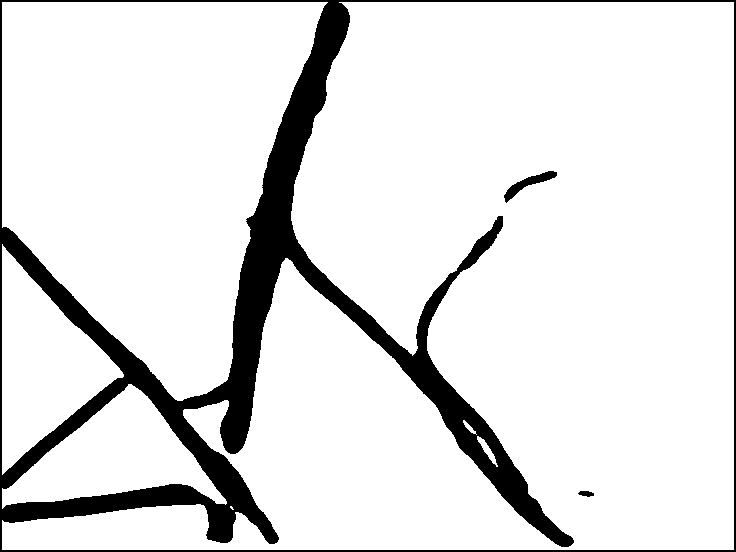}}
    \hfill
  \subfloat{%
        \includegraphics[width=0.19\linewidth]{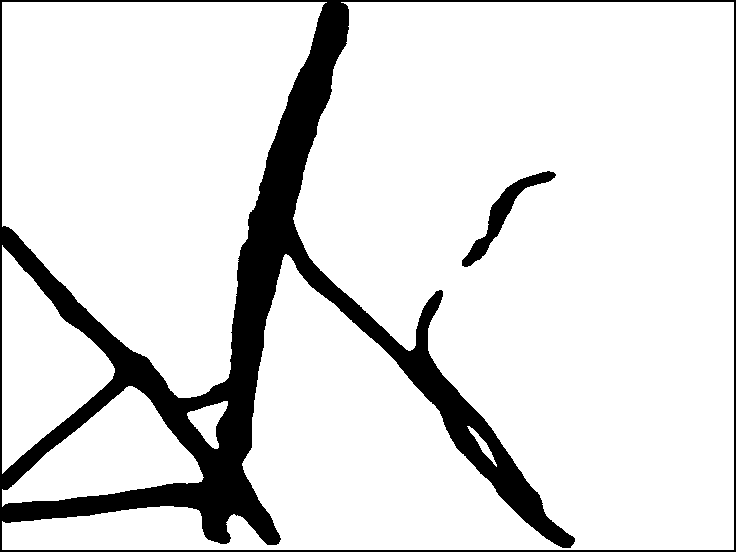}}
    \\
  \subfloat{%
        \includegraphics[width=0.19\linewidth]{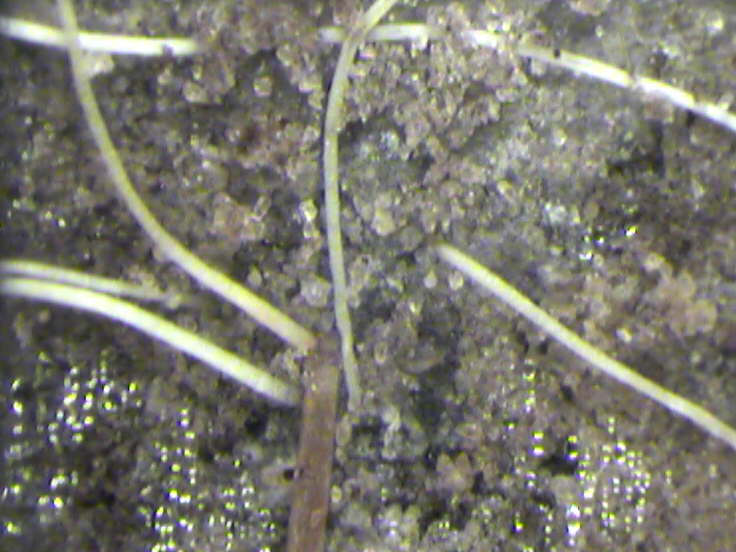}}
    \hfill
  \subfloat{%
        \includegraphics[width=0.19\linewidth]{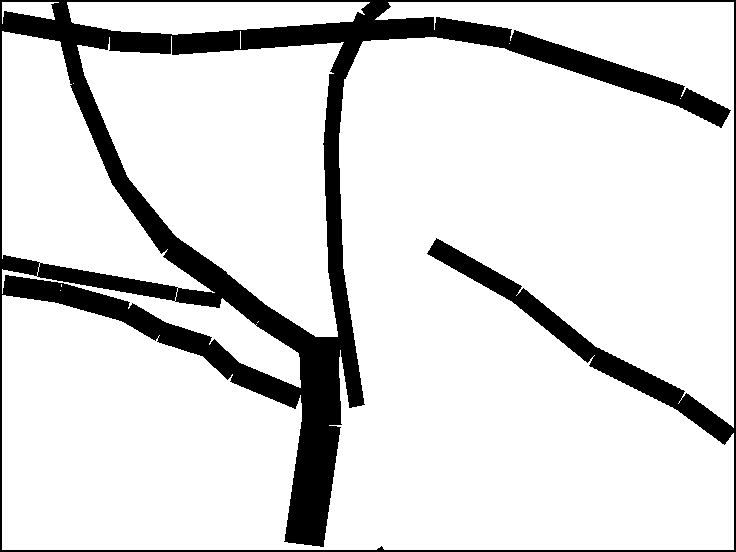}}
    \hfill
  \subfloat{%
        \includegraphics[width=0.19\linewidth]{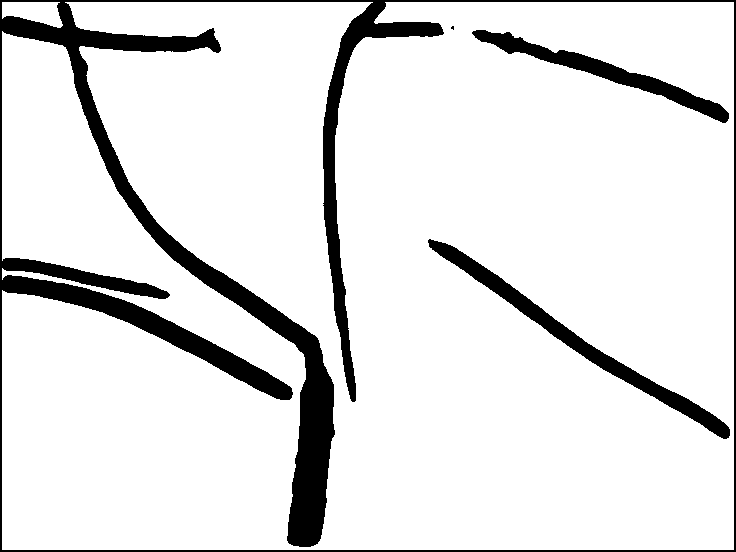}}
    \hfill
  \subfloat{%
        \includegraphics[width=0.19\linewidth]{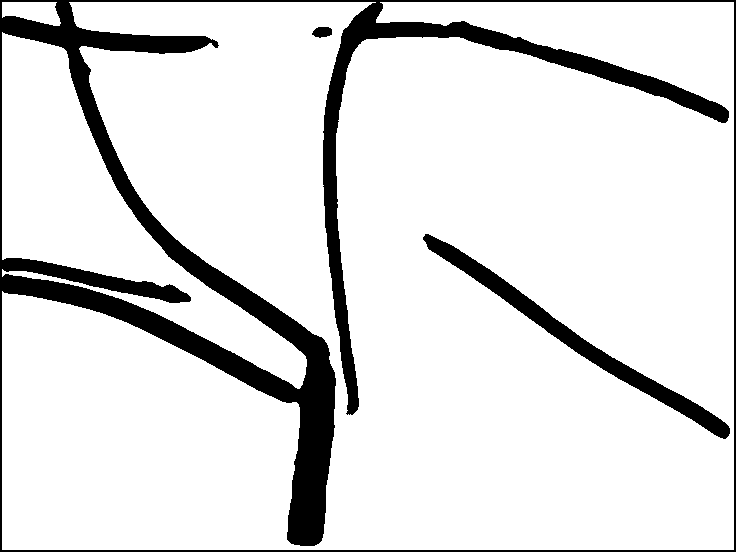}}
    \hfill
  \subfloat{%
        \includegraphics[width=0.19\linewidth]{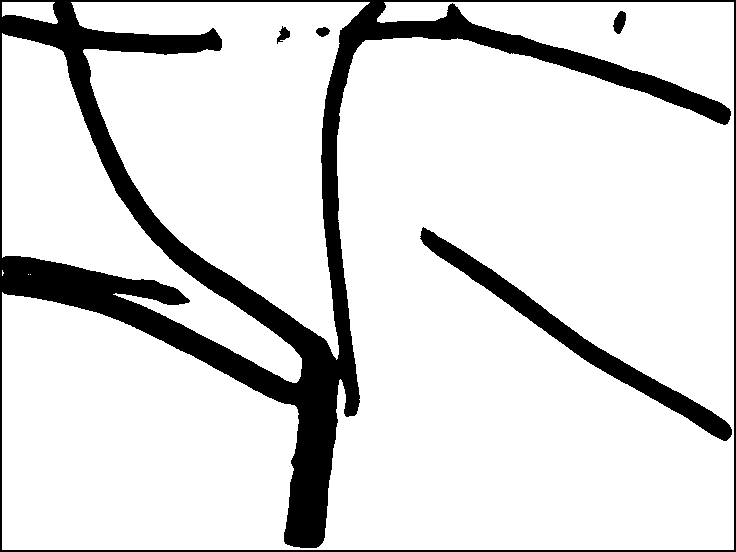}}
    \\
  \subfloat{%
        \includegraphics[width=0.19\linewidth]{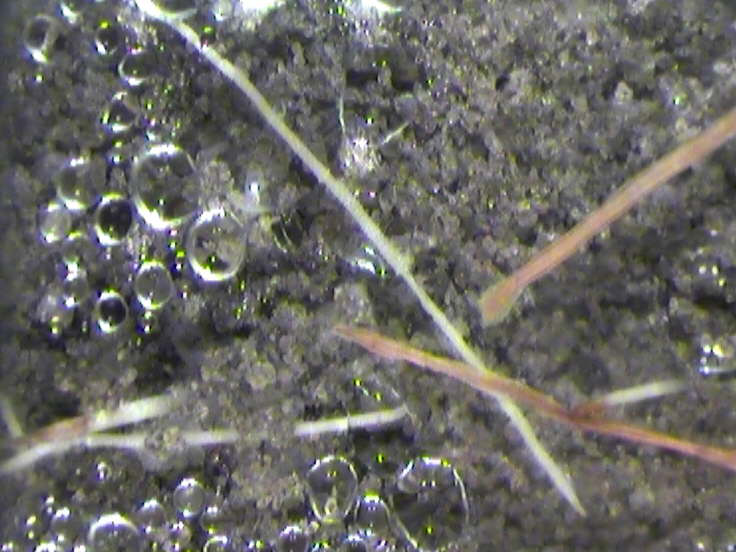}}
    \hfill
  \subfloat{%
        \includegraphics[width=0.19\linewidth]{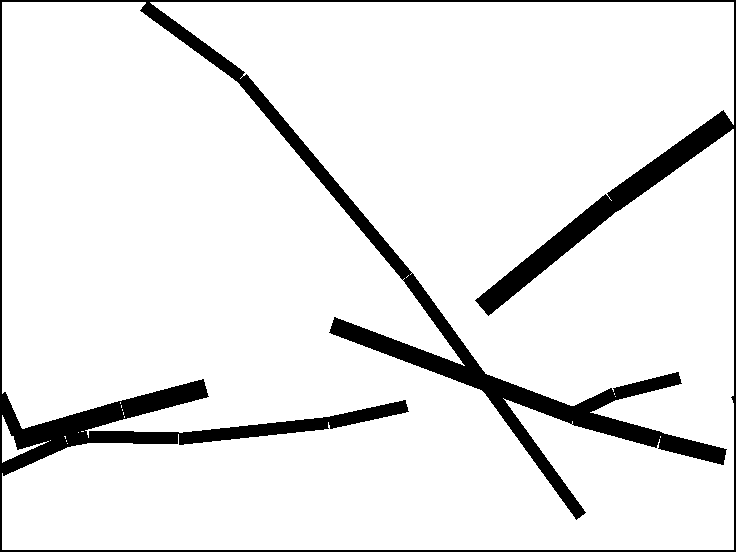}}
    \hfill
  \subfloat{%
        \includegraphics[width=0.19\linewidth]{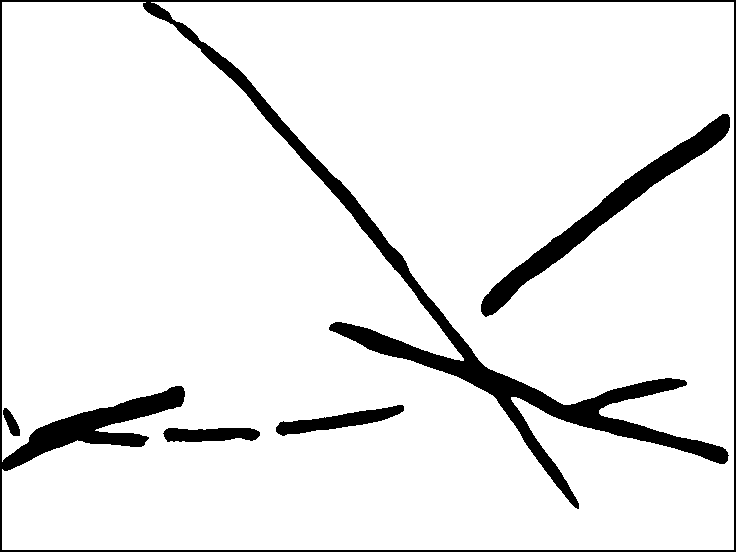}}
    \hfill
  \subfloat{%
        \includegraphics[width=0.19\linewidth]{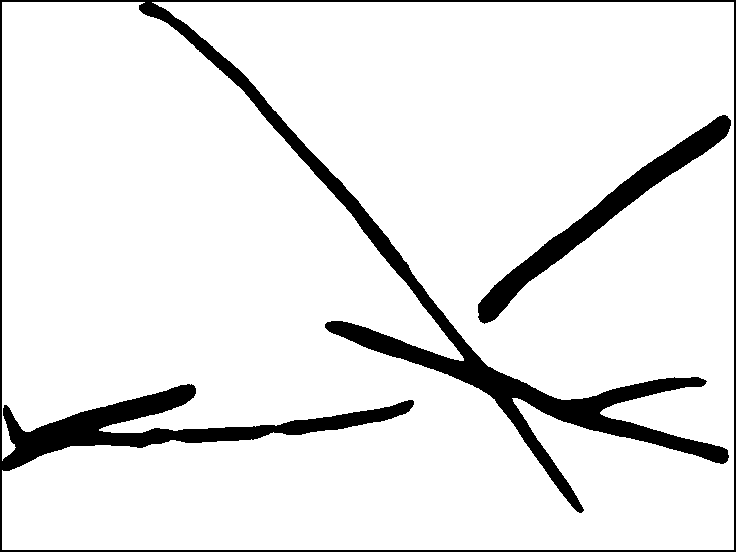}}
    \hfill
  \subfloat{%
        \includegraphics[width=0.19\linewidth]{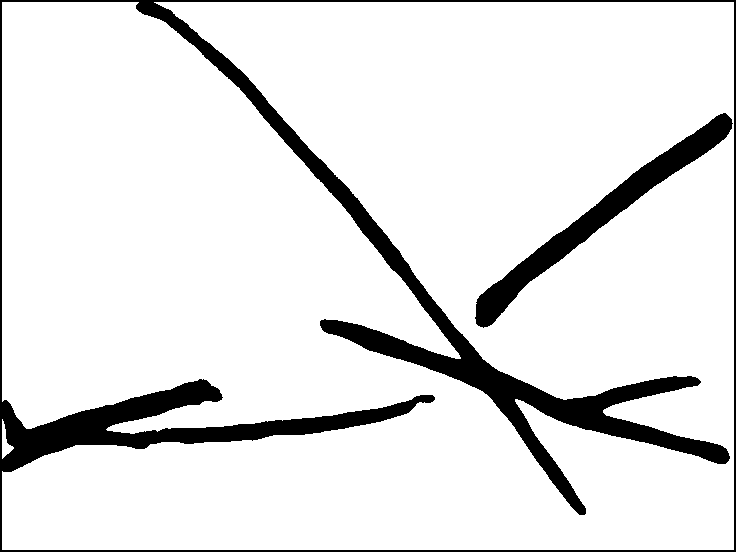}}
    \\
  \subfloat{%
        \includegraphics[width=0.19\linewidth]{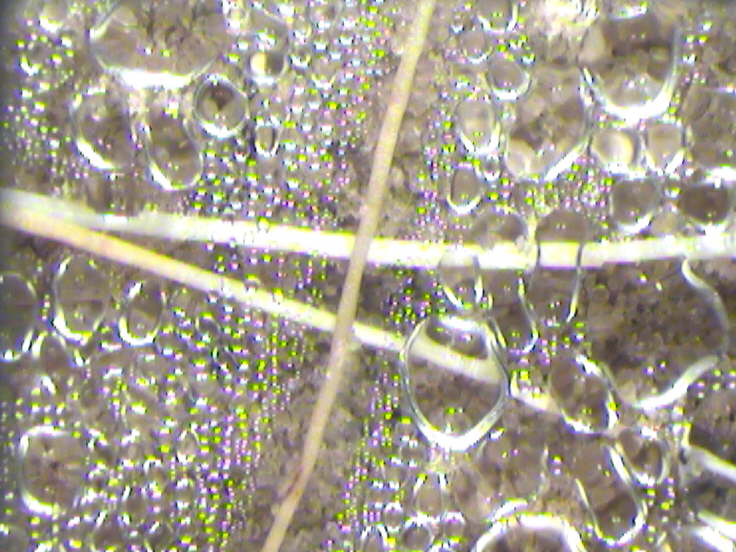}}
    \hfill
  \subfloat{%
        \includegraphics[width=0.19\linewidth]{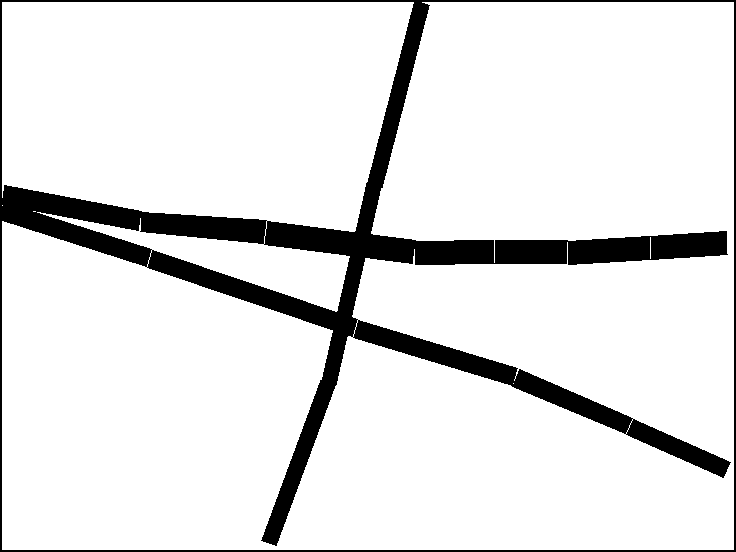}}
    \hfill
  \subfloat{%
        \includegraphics[width=0.19\linewidth]{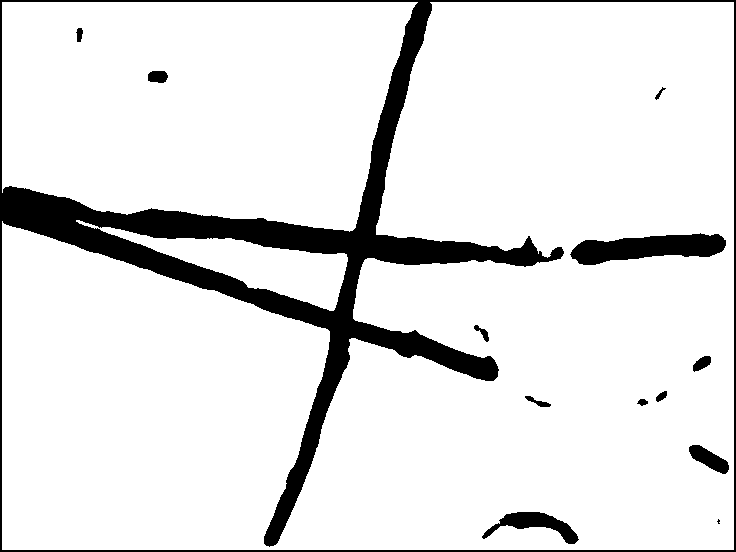}}
    \hfill
  \subfloat{%
        \includegraphics[width=0.19\linewidth]{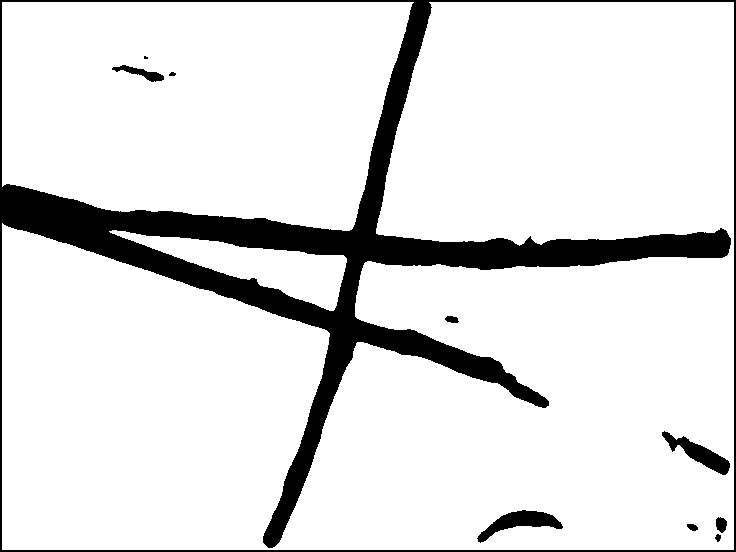}}
    \hfill
  \subfloat{%
        \includegraphics[width=0.19\linewidth]{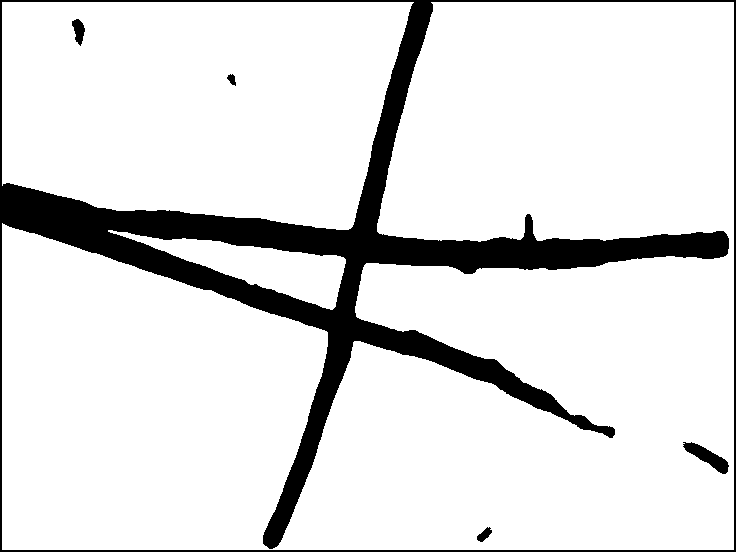}}
\end{center}
  \caption{Segmentation results of selected test images taken across differing depths, dates, and local environments. Column (a) shows raw input peanut root minirhizotron images. Column (b) shows the manually labeled ground truth masks. Columns (c)-(e) show the segmentation results of models with depth 4, 5 and 6, respectively.}
  \label{SegPeanut} 
\end{figure*}

\textbf{Experimental results}.
Selected examples of segmentation results of the three models with different depths are shown in Figure \ref{SegPeanut}.
Column (a) shows the raw peanut root images taken from the minirhizotron system.
These images were taken across differing depths, dates, and local environments.
The corresponding manually labeled ground truth masks are shown in column (b).
Column (c)-(e), show segmentation masks of our models with depth 4, depth 5 and depth 6, respectively.
Global threshold was set to be 0.4 to generate the binary segmentation mask.
Qualitatively, all three models provided good segmentation results.
Most of the roots can be segmented from complicated soil backgrounds.
Our method managed to correct two major defects caused by the manual annotation process using WinRHIZO software: (1) fixed label width along rectangles; and (2) gaps between neighboring rectangles.
Our model can capture the real thickness and diameter variation along each root.
The segmented roots are consistent and smooth instead of having a boxy shape with gaps in manually labeled ground truth masks.
As shown in column (c)-(e), our segmentation masks can generate better binary masks of roots, which can help with accurate determination of root traits in subsequent measurements, such as length, diameter and surface area, etc.

Some interesting details were observed in the segmentation results in the last three rows in Figure \ref{SegPeanut}.
In the third row, part of the top of the root is covered by soil in the raw peanut root image. 
An example of the robustness of our method is that a small area of the ground truth mask was mislabeled; however, our method obtained the correct answer in all three models. 
In the fourth row, the root at the bottom left of the picture is partially covered by soil.
We expected that it would be considered as a single piece of root as it was labeled as such in the ground truth mask.
The shallow model (depth 4) generated three separated small pieces of root instead of one unbroken piece.
In contrast, deeper models (depth 5 and depth 6) were capable of filling the gap and generated an unbroken root.
This ability is very important when considering the density or number of roots in a specific area.
The last row shows the case of a very complicated background.
Because of a wide variety of reflections, it is very difficult to eliminate water bubbles in segmentation results.
Our method was able to remove most of the water bubbles, but there was still some residual noise in the segmentation results.
The output masks from depth 6 model are much cleaner than depth 5 and depth 4 models, which indicates that the deeper network is more powerful to accommodate complex noise in order to match ground truth masks as close as possible.
This seems reasonable because a deeper network can possibly extract higher-level features to further improve the reconstruction step in decoders.

\begin{figure}[t]
\begin{center}
  \subfloat{%
       \includegraphics[width=1.0\linewidth]{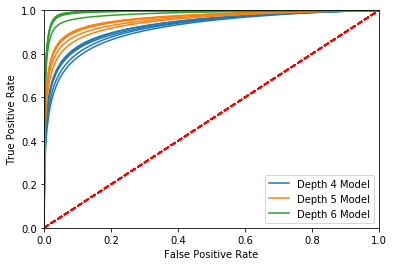}}
\end{center}
  \caption{ROC curves plotted for models with depth 4 (Blue), depth 5 (Orange), and depth 6 (Green), respectively. Each model was trained five times with random weight initialization.}
  \label{ROC} 
\end{figure}

\begin{table}[t]
\renewcommand{\arraystretch}{1.3}
\caption{Average AUC and standard deviation of AUC calculated from ROC curves for different models}
\label{tab:PeanutAUC}
\begin{center}
\begin{tabular}{c|c}
    \hline
    Model & Average AUC $\pm$ Std. AUC\\
    \hline
    \hline

    Depth 4 & 0.9259 $\pm$ 0.007 \\
    \hline
    
    Depth 5 & 0.9565 $\pm$ 0.007 \\
    \hline
    
    Depth 6 & 0.9904 $\pm$ 0.005 \\
    \hline
\end{tabular}
\end{center}
\end{table}

In order to evaluate the consistency of the models, we trained each model 100 epochs for five trials with different random weight initialization.
We calculate the TPR and FPR using the entire test dataset containing 0.7 billion pixels.
The ROC curves for each model are shown in Figure \ref{ROC}.
The average and Std. of AUC for each model are shown in Table \ref{tab:PeanutAUC}.
The depth 6 model had the highest average AUC of 0.9904 indicating the best segmentation accuracy among all the models.
Additionally the method showed good consistency as all three models had small variance in AUC.

\subsection{Transfer Learning on Limited Switchgrass Root Dataset}
\textbf{Experiment setup}.
For specific applications such as plant root segmentation, the dataset is often limited like our switchgrass root dataset. Transfer learning is a promising technique to help with training. 
In our experiments, we had two available datasets to get pre-trained features: the pre-trained features from popular massive-scale ImageNet dataset and our self-collected peanut root dataset.
Our situation is similar to most data-limited applications where we often have some moderate-scale datasets that relevant to the target dataset.
We designed experiments on the small-scale switchgrass root dataset to explore the effect of pre-trained features from these two different datasets.
Compared with the ImageNet dataset that has 14 million images, our peanut root dataset is quite small, but much more relevant to the switchgrass root dataset. 
As the goal in this experiment is to better understand pre-trained features on a general model instead of finding the highest performing network architecture, we implemented the U-net based model with down path architecture the same as the VGG13 network \citep{simonyan2014very}.
Since ImageNet features were trained from a classification network, we could only take all the features in convolutional layers from the VGG13 network and apply them to our encoder as the pre-trained features in our model.
Unlike the ImageNet dataset, our peanut dataset had pixel-level labels, which could be used to directly train our U-net based model for segmentation.
An advantage of the pre-trained features using the peanut dataset is that we could compute the pre-trained features for both encoder and decoder.
We believe those features in decoder are also crucial for improving segmentation performance, because decoder blocks also extract higher-level feature maps for up-sampling.
Thus, we studied pre-trained features not only in encoder, but also in the combination of encoder and decoder. To make a comprehensive comparison of different pre-trained features, we implemented four models namely: 1) S-model whose weights are randomly initialized; 
2) I-model whose encoder is initialized with pre-trained weights on ImageNet dataset;
3) P-En-model whose encoder is initialized with pre-trained weights on our peanut dataset; and
4) P-EnDe-model whose encoder as well as decoder are initialized with pre-trained weights on our peanut dataset.
These four models have exactly the same architecture but different weight initialization.
The S-model works as base-line model in the comparison.
All the models were trained on the switchgrass dataset for 300 epochs.
We used a relatively larger learning rate $5e^-5$ for the S-model, because it was trained from scratch and a smaller learning rate of $1e^-5$ for the I-model, P-En-model and P-EnDe-model.
Since randomly initialized weights can cause variance in segmentation results, each model was trained five times to compare the performance consistency.
To make sure the evaluation is accurate, we used 7 well-annotated switchgrass minirhizotron images for evaluating the performance of each model and the remaining 21 images for training.
Due to the limitation of GPU memory, each switchgrass root image was evenly cropped into 15 small images with size 720x510 pixels. 
Unlike the peanut root dataset, the number of root pixels versus non-root pixels is heavily imbalanced in our switchgrass root imagery. We calculated the ratio of non-root pixels and root pixels for each cropped image. This ratio varies significantly across these images. We set a positive weight of 20 (close to the median ratio value) for root pixels in the binary cross-entropy loss function to prevent models from being heavily biasing towards the large number of non-root pixels. In addition, it is difficult to evaluate the classification accuracy of the root class on switchgrass root images based on ROC curves, because the overall classification accuracy will be overwhelmed by the classification accuracy for non-root pixels.
Therefore, we plotted the PR curves to highlight the classification performance of root pixels specifically to show how the different pre-trained features can help with the classification of root pixels.

\begin{figure}[h]
\begin{center}
  \subfloat[\label{PRC_E0-300}]{%
       \includegraphics[width=0.98\linewidth]{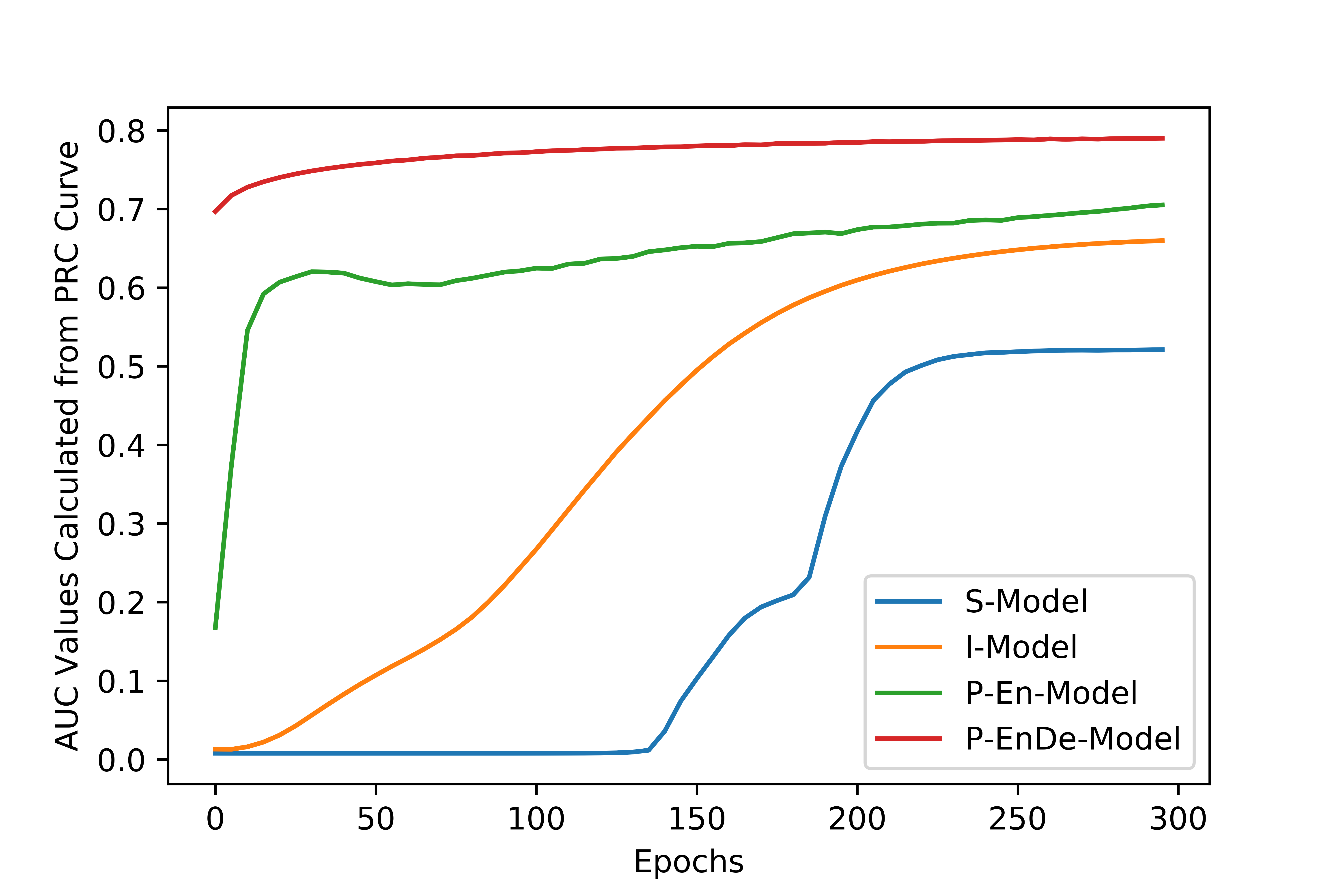}}
    \hfill
  \subfloat[\label{PR_E300}]{%
        \includegraphics[width=0.98\linewidth]{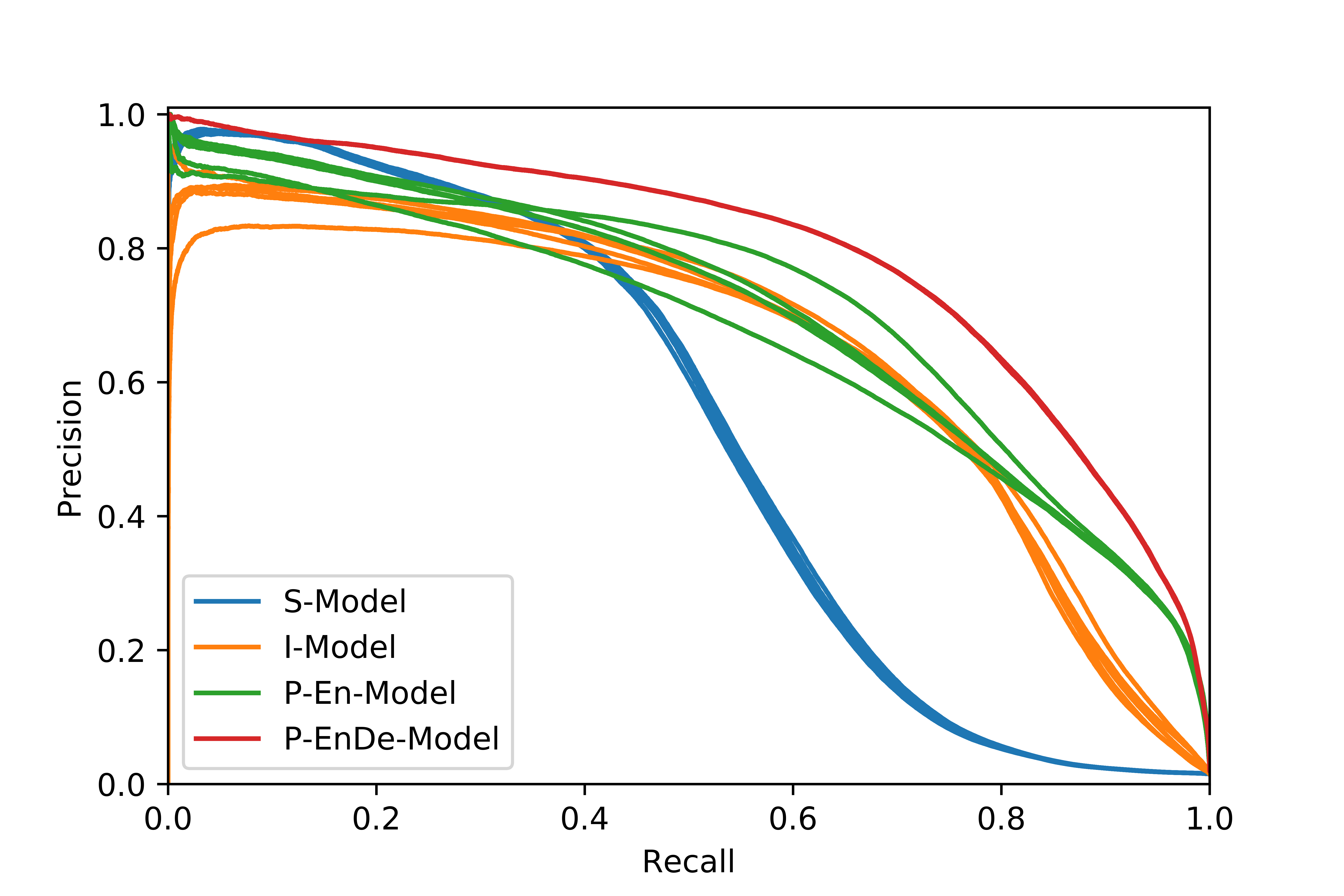}}
    \hfill
\end{center}
  \caption{(a) AUC values for PR curves curves at each training epoch for all the models with different weight initialization. Blue curve: S-model, weights in both encoder and decoder are randomly initialized. Orange curve: I-model, weights in encoder are initialized with ImageNet pre-trained features and randomly initialized weights in decoder. Green curve: P-En-model, weights in encoder are initialized with peanut pre-trained features and randomly initialized weights in decoder. Red curve: P-EnDe-model, weights in both encoder and decoder are initialized with peanut pre-trained features. (b) PR curves for S-model, I-model, P-En-model and P-EnDe-model at epoch 300.}
  \label{Trans_PR} 
\end{figure}

\begin{figure*}[p]
   \centering
       \includegraphics[page=1,width=1.0\textwidth]{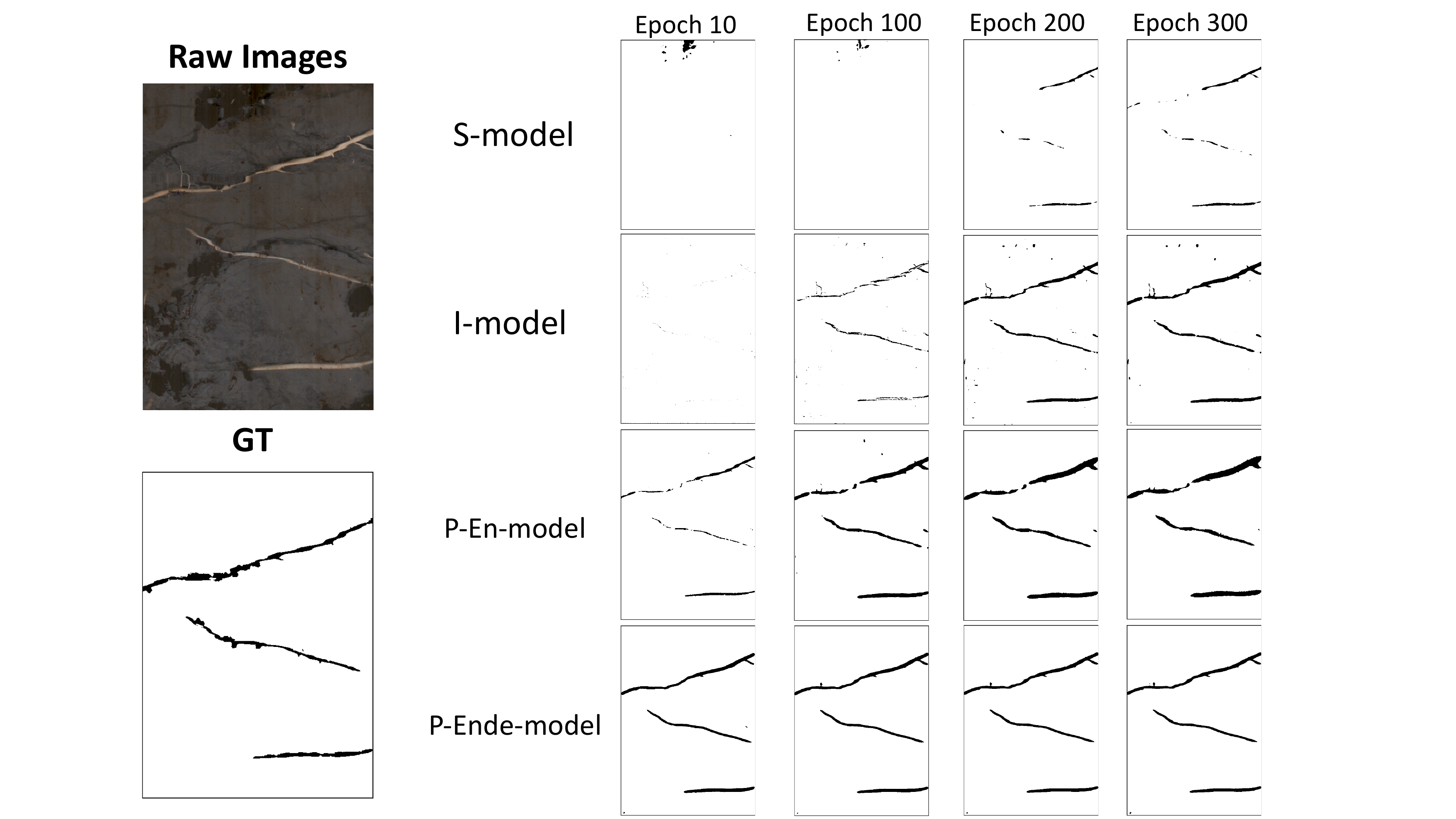}
    \hfill
      \includegraphics[page=1,width=1.0\textwidth]{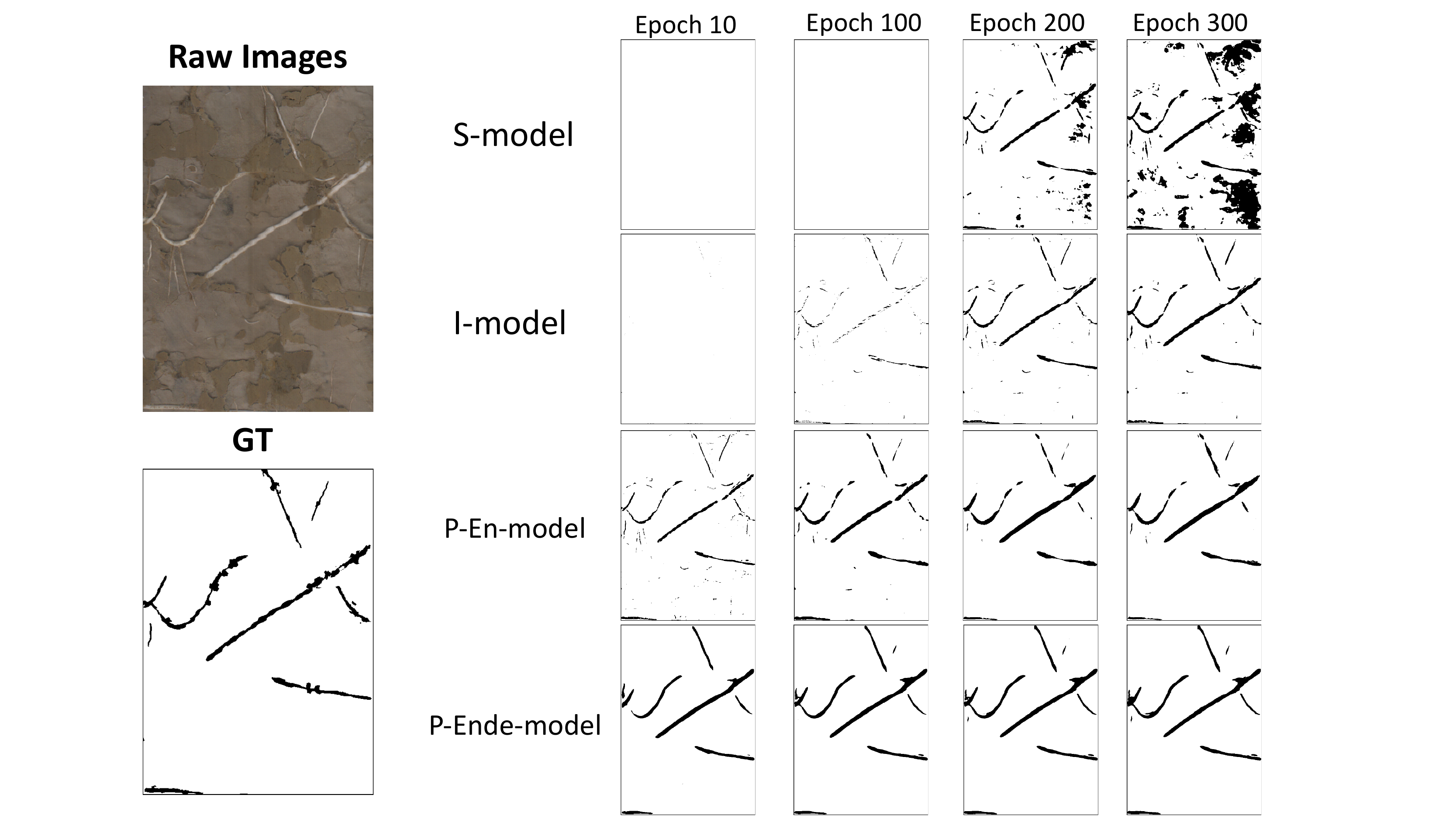} \\
 \caption{Segmentation results of selected test images generated by S-model, I-model, P-En-model and P-EnDe-model at epoch 10, 100, 200 and 300. }
 \label{fig:TransSeg}
\end{figure*}

\textbf{Experimental results}. 
We selected two switchgrass root images with the best manually labeled ground truth to show binary segmentation masks generated by one of each of the four models at epoch 10, 100, 200 and 300 as shown in Figure \ref{fig:TransSeg}.
The quality of the binary segmentation mask is sensitive to probability thresholds which usually vary for different models.
To make a fair comparison, we plotted ROC curves for each model at different epochs and picked threshold corresponding to the FPR value at 1\%.
The FPR value was calculated based on all test images, but it could vary for each individual image.
Qualitatively, the S-model could not classify roots until several hundreds epochs due to the limited training data. 
As shown in Figure \ref{fig:TransSeg} S-model failed to segment tiny or fine roots and mis-classified (false positive) a lot of non-root pixels when the color of soil background was similar to roots. 
The I-model performed a bit better than the S-model, but not as good as the P-En-model and P-EnDe-model.
There were still some noise mis-classified as root pixels in the background.
The segmented images were much cleaner for P-En-model and P-EnDe-model.
Most importantly, starting from the early training epochs (e.g. epoch 10), the P-En-model and P-EnDe-model already had decent segmentation results, indicating that peanut pre-trained features were more effective than ImageNet pre-trained features.

\begin{figure}[t!]
\begin{center}
  \subfloat[\label{ROC_E0-300}]{%
       \includegraphics[width=0.98\linewidth]{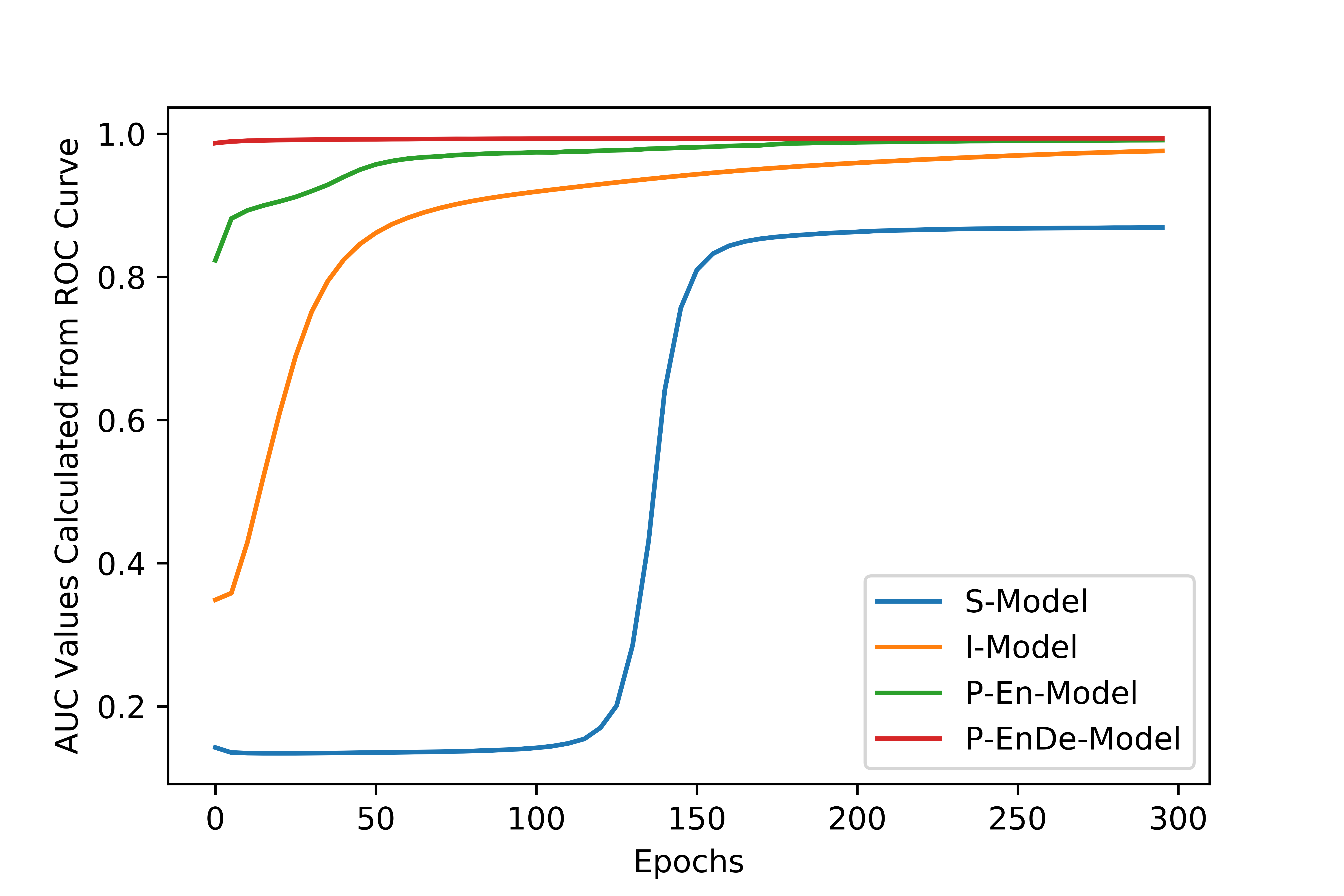}}
    \hfill
  \subfloat[\label{ROC_E300}]{%
       \includegraphics[width=0.98\linewidth]{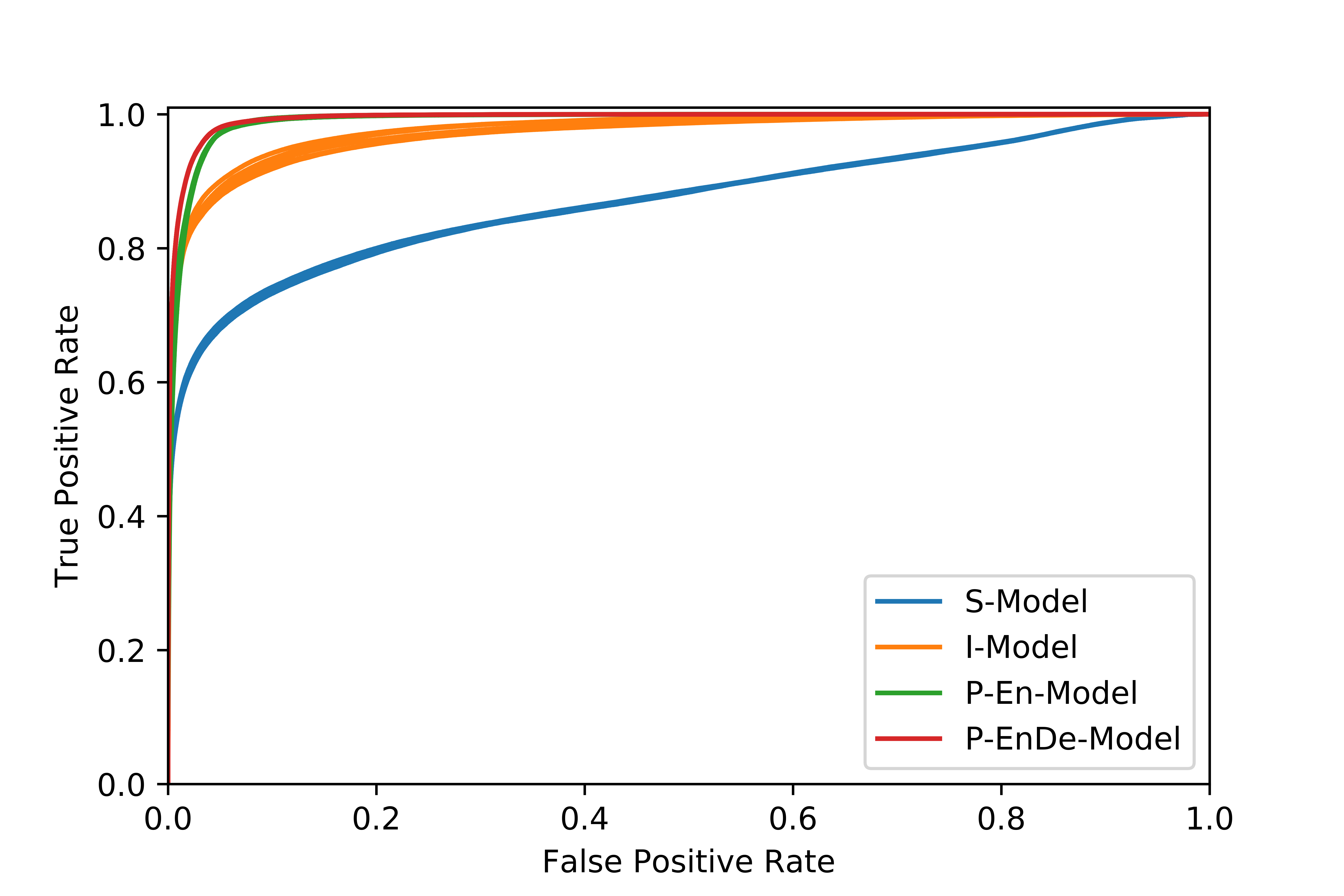}}
\end{center}
  \caption{(a) AUC values for ROC curves curves at each training epoch for all the models with different weight initialization. Blue curve: S-model, weights in both encoder and decoder are randomly initialized. Orange curve: I-model, weights in encoder are initialized with ImageNet pre-trained features and randomly initialized weights in decoder. Green curve: P-En-model, weights in encoder are initialized with peanut pre-trained features and randomly initialized weights in decoder. Red curve: P-EnDe-model, weights in both encoder and decoder are initialized with peanut pre-trained features. (b) ROC curves for S-model, I-model, P-En-model and P-EnDe-model at epoch 300.}
  \label{Trans_ROC} 
\end{figure}

To evaluate the overall segmentation performance for each model, we calculated AUC values based on the ROC curves for all models at each training epoch as shown in Figure \ref{ROC_E0-300}. We also showed the ROC curves for each model (5 trials for each model) at the final epoch (300) as shown in Figure \ref{ROC_E300}.
AUC values indicate the segmentation accuracy on both root pixels and soil pixels.
Starting from the very beginning, models with pre-trained features had higher AUC values than the S-model that was trained from scratch. 
Under the same training condition, pre-trained features improved the overall segmentation performance substantially.
Specifically, features pre-trained on the peanut root dataset were more effective than features pre-trained on ImageNet dataset, even though the ImageNet dataset is much larger than the peanut dataset.

To quantitatively analyze the effectiveness of pre-trained features, we showed AUC values calculated from PR curves for all models at each training epoch and the PR curves at the 300 epoch for 5 trials in Figure \ref{Trans_PR}.
PR curves show classification performance on root class. 
High AUC values in PR curves represent both high precision and recall values, indicating that features are more effective to classify root pixels.
P-En-model and P-EnDe-model had higher AUC values, which illustrates that peanut pre-trained features are more helpful than ImageNet pre-trained features for switchgrass root segmentation.
Additionally, models with pre-trained features converged faster, especially the P-EnDe-model with both pre-trained encoder and decoder on peanut dataset. 

The average and standard deviation value of the AUC among those 5 trials are shown in Table \ref{tab:TransAUC}.
According to the ROC curves, the pre-trained features could improve segmentation performance a lot.
The standard deviation values of the AUC showed the consistency of the model performance with different weight initialization.
Models used with the pre-trained encoders had relatively large variance compared with S-model trained from scratch, which showed that the segmentation performance was more sensitive to the randomly weight initialization in decoder.
In terms of the overall segmentation accuracy, models used with the peanut-pre-trained encoder were more consistent (lower standard deviation of AUC on ROC curve) than models used with the ImageNet encoder.
P-EnDe-model had the best stability due to its smallest std. AUC values on both ROC and PR curves.
The S-model had a poor precision score when recall was close to 1, which indicates that a comparable amount of soil pixels were mis-classified as root.
Furthermore, the P-En-model and P-EnDe-model with peanut pre-trained features offered much higher precision scores when recall was close to 1, which proves that the peanut features were more relevant to roots instead of soil.
Also, the P-EnDe-model had the highest average AUC values on both ROC and PR curves indicating that pre-trained features in the decoder also were important for segmentation tasks.

\begin{table}
\renewcommand{\arraystretch}{1.3}
\caption{Average AUC and standard deviation of AUC calculated from ROC and PR curves for different models}
\label{tab:TransAUC}
\begin{center}
\begin{tabular}{c|c|c}
    \hline
    Model & \multicolumn{2}{c}{Average AUC $\pm$ Std. AUC} \\
    \hline
    {} & ROC Curves & PR Curves \\
    \hline
    \hline
    S-model & 0.870 $\pm$ 0.001 & 0.425 $\pm$ 0.003 \\
    \hline
    I-model & 0.973 $\pm$ 0.004 & 0.652 $\pm$ 0.008 \\
    \hline
    P-En-model & 0.991 $\pm$ 0.000$^a$ & 0.700 $\pm$ 0.020 \\
    \hline
    P-EnDe-model & \textbf{0.994 $\pm$ 0.000$^b$} & \textbf{0.790 $\pm$ 0.000$^c$}\\
    \hline
\end{tabular}
\end{center}
\footnotesize{Only three significant digits are shown in the table. The actual value is: $^a$ 0.0004, $^b$ 0.00005 and $^c$ 0.0004}
\end{table}

\section{Discussion}
In our experiments, we explored the effectiveness of pre-trained features from the popular ImageNet dataset and a peanut root dataset for switchgrass root segmentation.
Comparing the results of all the models, pre-trained features from both datasets can not only improve the segmentation performance, but also help a model converge faster and thus save a large amount of training time.
Although the ImageNet dataset is massive in scale, it appears that pre-trained features from peanut dataset had better performance.
Thus, (perhaps, intuitively) pre-trained features from the massive-scale ImageNet dataset are not always the best for imagery with different visual appearances such as plant root images.
The relevance of the pre-trained dataset with respect to the target dataset is more crucial for segmentation performance.
Since the ImageNet dataset and the switchgrass dataset are very different, it is likely that only the features in shallow layers can help with segmentation results, since low-level features are more general such as edges or textures \citep{yosinski2014transferable}.
The higher-level features are more likely to be problem-specific, which could mislead the decision of a model on the target dataset.
This may be even more pronounced when the model is deeper, because the proportion of effective parameters for our application in shallow layers is getting smaller.
In contrast, features from the peanut root dataset that is much smaller but highly related to the switchgrass dataset are more valuable regardless of how deep the model is, because both low-level and high-level features are useful to the switchgrass root images.

It is promising to potentially combine our model with existing annotation tools (rhizoTrak, TrakEM2) to save extensive annotation effort for analysis of plant root minirhizotron images. 
Our model can pre-segment all raw images in a first step to get annotation for each pixel. 
Then, these pre-segmented images can be passed to rhizoTrak to correct mis-labeled parts by technicians, which can save a lot of effort for manual annotation process.
With the help of pre-trained features from other species, our model can generate pixel-level annotation with acceptable quality even though the target dataset is limited.

\section{Conclusions}
In this work, we proposed the use of U-net based deep neural networks for automated, precise, pixel-wise segmentation of plant roots in minirhizotron imagery.
Our model achieved high quality segmentation masks with 99.04\% ROC AUC at the pixel-level and overcame errors in human-labeled ground truth masks on peanut root dataset.
We also found that deep networks can better resolve more challenging images (more complicated backgrounds) than shallow networks.
Furthermore, we improved the segmentation performance on a small-scale switchgrass root dataset by using pre-trained features from the massive-scale ImageNet dataset and a mid-scale peanut root dataset, followed by fine tuning on a small switchgrass root dataset.
We obtained above 99\% segmentation accuracy in switchgrass root segmentation with pre-trained encoder and decoder from our peanut root dataset.
Our results indicate that both pre-trained encoder and decoder can help with segmentation performance when the target dataset is small.
The pre-trained features can help a model converge faster and have much more stable performance.
Also, features pre-trained on peanut dataset that is relatively small but highly related to the switchgrass dataset were more effective than those pre-trained on a massive-scale but less relevant ImageNet dataset.

\section*{Acknowledgement}

This work was partially supported by U.S. Department of Energy, Office of Science, Office of Biological and Environmental Research award number DE-SC0014156. The information, data, or work presented herein was partially funded in part by the Advanced Research Projects Agency-Energy (ARPA-E), U.S. Department of Energy, under Award Number DE-AR0000820. The views and opinions of authors expressed herein do not necessarily state or reflect those of the United States Government or any agency thereof.

\newpage
\bibliographystyle{elsarticle-harv} 

\end{document}